\documentclass[12pt]{wlscirep}
\usepackage[utf8]{inputenc}
\usepackage{float}
\usepackage[T1]{fontenc}
\usepackage[draft]{changes}

\usepackage{amsmath,amsfonts,bm}

\def\eqref#1{equation~\ref{#1}}

\def\1{\bm{1}}

\def\ve{{\bm{e}}}
\def\vf{{\bm{f}}}

\def\vh{{\bm{h}}}

\def\vu{{\bm{u}}}
\def\vv{{\bm{v}}}

\def\vz{{\bm{z}}}
\def\vepsilon{{\bm{\epsilon}}}
\def\valpha{{\bm{\alpha}}}
\def\vomega{{\bm{\omega}}}

\def\mA{{\bm{A}}}
\def\mB{{\bm{B}}}

\def\mD{{\bm{D}}}

\def\mF{{\bm{F}}}

\def\mH{{\bm{H}}}
\def\mI{{\bm{I}}}

\def\mK{{\bm{K}}}
\def\mL{{\bm{L}}}
\def\mM{{\bm{M}}}

\def\mP{{\bm{P}}}
\def\mQ{{\bm{Q}}}
\def\mR{{\bm{R}}}

\def\mU{{\bm{U}}}
\def\mV{{\bm{V}}}
\def\mW{{\bm{W}}}

\def\mZ{{\bm{Z}}}

\DeclareMathAlphabet{\mathsfit}{\encodingdefault}{\sfdefault}{m}{sl}
\SetMathAlphabet{\mathsfit}{bold}{\encodingdefault}{\sfdefault}{bx}{n}

\def\gA{{\mathcal{A}}}

\def\gE{{\mathcal{E}}}

\def\gG{{\mathcal{G}}}

\def\gL{{\mathcal{L}}}

\def\gN{{\mathcal{N}}}

\def\sR{{\mathbb{R}}}

\newcommand{\E}{\mathbb{E}}

\usepackage{graphicx}%
\usepackage{multirow}%
\usepackage{mathrsfs}%

\usepackage{algorithm}
\usepackage{algorithmic}

\usepackage[title]{appendix}%
\usepackage{xcolor}%
\usepackage{textcomp}%
\usepackage{manyfoot}%
\usepackage{booktabs}%
\usepackage{listings}%
\usepackage{lineno}
\usepackage[square,sort,comma,numbers]{natbib}
\usepackage{listings}

\definecolor{codegreen}{rgb}{0,0.6,0}
\definecolor{codegray}{rgb}{0.5,0.5,0.5}
\definecolor{codepurple}{rgb}{0.58,0,0.82}
\definecolor{backcolour}{rgb}{0.95,0.95,0.92}

\lstdefinestyle{mystyle}{
    backgroundcolor=\color{backcolour},   
    commentstyle=\color{codegreen},
    keywordstyle=\color{magenta},
    numberstyle=\tiny\color{codegray},
    stringstyle=\color{codepurple},
    basicstyle=\ttfamily\footnotesize,
    breakatwhitespace=false,         
    breaklines=true,                 
    captionpos=b,                    
    keepspaces=true,                 
    showspaces=false,                
    showstringspaces=false,
    showtabs=false,                  
    tabsize=2
}

\usepackage{hyperref}
\hypersetup{
    linktoc=page  
}

\usepackage{zref-user}
\usepackage{xurl}

\usepackage{bibunits}
\usepackage{thm-restate}
\usepackage{multirow}
\usepackage{colortbl}
\usepackage{titletoc} 
\usepackage{etoolbox}

\renewcommand{\figurename}{Fig.}
\makeatletter
\newcommand{\settitle}{\@maketitle} 
\makeatother
\DeclareCaptionType{supfigure}[Supplementary Fig.][List of Supplementary Figures]
\DeclareCaptionType{suptable}[Supplementary Table][List of Supplementary Tables]

\makeatletter
\patchcmd{\listof}
  {\float@listhead}
  {\@namedef{l@#1}{\l@supfigure}\float@listhead}
  {}{}
\makeatother

\begin{document}

\title{An Equivariant Pretrained Transformer for Unified 3D Molecular Representation Learning}

\author[1,2,+]{Rui Jiao}
\author[1,2,+]{Xiangzhe Kong}
\author[1,2,+]{Li Zhang}
\author[1,2]{Ziyang Yu}
\author[1,2,4]{Fangyuan Ren}
\author[1,2]{Wenjuan Tan}
\author[3,*]{Wenbing Huang}
\author[1,2,*]{Yang Liu}

\affil[1]{Department of Computer Science and Technology, Tsinghua University, Beijing, China}
\affil[2]{Institute for AI Industry Research, Tsinghua University, Beijing, China}
\affil[3]{Gaoling School of Artificial Intelligence, Renmin University of China, Beijing, China}
\affil[4]{Department of Pharmacy, University of Pisa, Pisa, Italy}
\affil[*]{Correspondence should be addressed to: hwenbing@ruc.edu.cn liuyang2011@tsinghua.edu.cn}

\affil[+]{These authors contributed equally to this work.}


\begin{abstract}
Pretraining on a large number of unlabeled 3D molecules has showcased superiority in various scientific applications. However, prior efforts typically focus on pretraining models in a specific domain, either proteins or small molecules, missing the opportunity to leverage cross-domain knowledge. To mitigate this gap, we introduce \textbf{E}quivariant \textbf{P}retrained \textbf{T}ransformer (EPT), 
an all-atom foundation model that can be pretrained from multiple domain 3D molecules. Built upon an E(3)-equivariant transformer, EPT is able to not only process atom-level information but also incorporate block-level features (\emph{e.g.} residuals in proteins). Additionally, we employ a block-level denoising task, rather than the conventional atom-level denoising, as the pretraining objective.
To pretrain EPT, we construct a large-scale dataset of 5.89M entries, comprising small molecules, proteins, protein-protein complexes, and protein-molecule complexes. Experimental evaluations on downstream tasks including ligand binding affinity prediction, protein property prediction, and molecular property prediction, show that EPT significantly outperforms previous state-of-the-art methods in the first task and achieves competitively superior performance for the remaining two tasks. Furthermore, we demonstrate the potential of EPT in identifying small molecule drug candidates targeting 3CL protease, a critical target in the replication of SARS-CoV-2. Among 1,978 FDA-approved drugs, EPT ranks 7 out of 8 known anti-COVID-19 drugs in the top 200, indicating the high recall of EPT. By using Molecular Dynamics (MD) simulations, EPT further discoveries 7 novel compounds whose binding affinities are higher than that of the top-ranked known anti-COVID-19 drug, showcasing its powerful capabilities in drug discovery.



\end{abstract}



\settitle

\begin{bibunit}


\section{Introduction}

Understanding and accurately representing the 3D geometric structure of molecular systems is critically important in numerous scientific domains, such as life sciences~\cite{eslami2022artificial}, drug discovery~\cite{blanco2023role}, and material design~\cite{pyzer2022accelerating}. This significance arises because 3D structures largely determine molecular properties in downstream tasks, which cannot be effectively captured by simpler 1D representations like SMILES for chemical molecules or amino acid sequences for proteins.
Recent advances in geometric Graph Neural Networks (GNNs)~\cite{schutt2018schnet, satorras2021n, liao2022equiformer} have enabled the accurate modeling of physical interactions between atoms while respecting the E(3) symmetry inherent in molecular systems. These models incorporate 3D coordinates for each node, ensuring that scalar attributes and dynamic processes remain invariant or equivariant under E(3) transformations. Strategies such as irreducible representations~\cite{thomas2018tensor}, frame averaging~\cite{puny2021frame}, and scalarization mechanisms~\cite{schutt2018schnet, satorras2021n} have been employed to preserve the E(3) symmetry. Geometric GNN models have demonstrated remarkable success in various applications, including molecular property prediction\cite{zhou2023uni}, protein generation~\cite{watson2023novo}, and antibody design~\cite{xu2022geodiff, hoogeboom2022equivariant, kong2023end}.


Despite remarkable successes, a central challenge remains: the scarcity of labeled data, which significantly limits the applicability of deep learning methods in scientific research.
To address this challenge, researchers have drawn inspiration from self-supervised pretraining techniques in Natural Language Processing (NLP), as exemplified by BERT~\cite{kenton2019bert} and GPT~\cite{radford2018improving}. Experimental results confirm that models pretrained on large-scale, unlabeled 3D molecular datasets exhibit substantial performance gains on downstream tasks after fine-tuning with limited labeled data. Specifically, for small molecules, approaches such as GraphMVP~\cite{liu2021pre} and 3D Infomax~\cite{stark20223d} apply contrastive learning from 2D-3D pairs, while MoleBlend~\cite{yu2023unified} aligns 2D and 3D features through a multimodal pretraining framework. For proteins, GearNet~\cite{zhang2022protein} uses contrastive learning on sequential and structural representations. Other works, such as Uni-Mol~\cite{zhou2023uni}, have explored cross-domain interactions, by pretraining separate models for small molecules and protein pockets before fine-tuning on binding datasets. In addition, inspired by score-based generative models, denoising tasks have emerged as powerful pretraining methods for learning force fields~\cite{zaidi2022pre, liu2022molecular, jiao2023energy, feng2023fractional}, with approaches like NERE~\cite{jin2023unsupervised} focusing on translation and rotation denoising at the ligand level. 

Existing pretraining methods predominantly focus on domain-specific models, such as those tailored to small molecules~\cite{luo2022one, jiao2023energy}, proteins~\cite{zhang2022protein}, or dual-tower frameworks that separately handle small molecules and proteins~\cite{zhou2023uni, feng2023protein}. However, these approaches are limited in their ability to generalize across multiple domains. We believe that developing a foundation model for diverse atomic systems is indispensable for advancing various scientific applications. On the one hand, recent breakthroughs in AI, such as the GPT series~\cite{radford2018improving, radford2019language, brown2020language} and Gato~\cite{reed2022generalist}, highlight the advantages of foundation models that unify learning across tasks and domains. These benefits include reduced need for handcrafted inductive biases, increased data diversity, and scalability with larger datasets and models. Extending this paradigm to molecular sciences could yield transformative results. On the other hand, from a physical perspective, all atomic systems, regardless of scale, are governed by the same fundamental principles, such as the Schrödinger Equation. Developing a single neural network capable of learning these universal principles represents a highly compelling pursuit.

Undoubtedly, building such a foundation model involves significant challenges that require addressing a range of technical and conceptual complexities. First, cross-domain data formats vary substantially. For small molecules, representations are typically at the atomic level, with each atom in a molecule corresponding to a specific set of features, such as atomic number, charge, and bond type. On the other hand, proteins are much more complex, requiring hierarchical representations where residues are composed of multiple atoms, and the structural context, including secondary structures, plays a critical role in determining function. This disparity in representation demands the development of a unified method capable of seamlessly integrating these different data formats while maintaining the necessary level of detail in each domain. Second, to capture the complex interactions between atoms and molecules, models must respect the fundamental symmetries that govern atomic systems. In particular, molecular interactions must be modeled while preserving E(3) symmetry, which accounts for the invariance of the system under rotations and translations in three-dimensional space. Finally, current self-supervised objectives used for pretraining are often task- or domain-specific. For example, objectives that work well for small molecules may not be directly applicable to proteins or other biological systems due to differences in their structural and functional properties. This limits the generalization capability of current models. 

In this paper, we propose Equivariant Pretrained Transformer (EPT),  an all-atom foundation model that is pretrained from multiple-domain 3D molecules, including small molecules, proteins, and complexes. This represents the first attempt to develop a general model to handle atomic systems of various types. 
By leveraging a unified molecular modeling component, EPT is able to not only process all-atom information but also incorporate block-level features that attend a broader context of each atom, such as the atom-level surroundings for small molecules and the residue-level belongings for proteins. More importantly, EPT conforms to E(3) symmetry and is thoroughly designed upon a generic transformer. It derives the embedding layer with one-layer equivariant GNN to reflect the graph geometry, and then update the atom-level scalar and vector features via the equivariant self-attention and feed-forward mechanisms in each layer. In addtion, EPT adopts a block-level denoised pretraining task, which requires the model to recognize the translation and rotation perturbations applied on each block, thereby enhancing the model's ability to model the complex hierarchical geometry of molecules. 

To pretrain EPT, we construct a large-scale pretraining dataset of over 5.89M entries by integrating existing datasets from different sources. 
We then demonstrate the efficacy of EPT across diverse downstream tasks and datasets, achieving state-of-the-art results in several cases. Specifically, for Ligand Binding Affinity (LBA) prediction, EPT outperforms existing methods on benchmarks with varying protein sequence similarities (30\% and 60\%). Furthermore, for the Mutation Stability Prediction (MSP) task, EPT showcases robust generalizability, significantly outperforming existing domain-specific models. For the Molecular Property Prediction (MPP) task, EPT achieves competitive results compared to leading approaches, with significant gains when using block-level denoising strategies. These results validate EPT's ability to unify geometric learning across domains, providing a robust and generalizable framework for 3D molecular modeling.
To further validate the utility of our EPT model, we apply EPT to rank FDA-approved drugs for their binding potential to the 3CL protease, a critical target in SARS-CoV-2 replication. The model successfully identifies known anti-COVID-19 drugs as top candidates, with further t-SNE analysis revealing that these drugs clustered closely with other high-affinity candidates predicted by EPT. From these clusters, we select 12 promising candidates for further validation through Molecular Dynamics (MD) simulations. Seven of these molecules exhibited higher binding affinities than the top-ranked known anti-COVID-19 drug. These results demonstrate the predictive power of EPT and its potential to accelerate drug discovery.

\begin{figure}[htbp]
    \centering
    \includegraphics[width=0.9\textwidth]{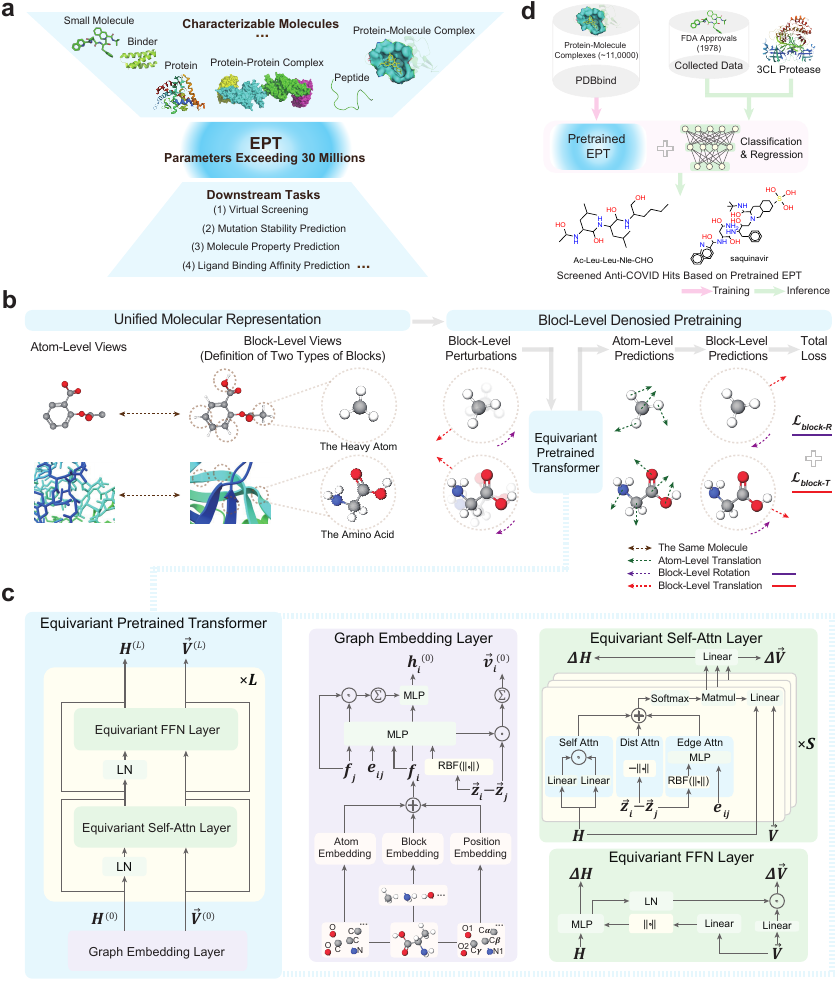}
    \caption{\textbf{\ \textbar \ Overview of EPT.} EPT is a foundation model for multi-domain 3D molecules.  
 \textbf{a} EPT is capable of addressing diverse downstream tasks, after pretrained on a large-scale hybrid dataset containing small molecules, proteins and complexes. \textbf{b}, EPT integrates molecules from different domains by defining "blocks" as the fundamental units for each domain. For small molecules, blocks are defined as heavy atoms and their associated hydrogens, while for proteins, blocks correspond to amino acids. During pretraining, blocks are perturbed by random translations and rotations around the center-of-mass (CoM), and EPT is trained to recover the original structure. \textbf{c}, In EPT, the atom representations including scalars $\mH$ and vectors $\mV$ are first initialized via a GNN-based embedding layer, and then updated by equivariant self-attention and feed-forward layers. \textbf{d}, We demonstrate the efficacy of the pretrained EPT model in virtual screening for anti-COVID-19 drugs, outperforming computational and learning-based baselines.}\label{fig1}
\end{figure}

%


\section{Results}

\subsection{Overview of EPT}
\label{sec:overview}

EPT is an all-atom foundation model for multiple domains including small molecules, proteins, and complexes (Fig.~\hyperref[fig1]{1a}). EPT creates a unified molecular representation across domains by defining "blocks" of atoms. For small molecules, each block is a collection of heavy atoms (non-hydrogen) and directly bonded hydrogens. For proteins, blocks are defined as amino acids. To enhance the model's ability to capture the block-level information, EPT utilizes a block-level pretraining strategy based on denoising (Fig.~\hyperref[fig1]{1b}). Specifically, each block is perturbed by a random translation of its center-of-mass (CoM) and a random rotation around the CoM. The model is trained to predict both the translational forces and angular momentum needed to restore the original structure, strengthening its ability to understand hierarchical geometries and maintain the physical integrity of molecular structures.

To effectively capture the geometry of molecular structures, EPT employs an improved Transformer architecture that integrates E(3) symmetry as its backbone model (Fig.\hyperref[fig1]{1c}). The process begins with a graph embedding layer that encodes both atom- and block-level information while initializing vector features through a message-passing layer. Subsequently, EPT alternates between equivariant self-attention and feed-forward layers to model interatomic relationships, producing rich scalar and vector representations. In total, EPT contains parameters exceeding 30M. Finally, the pretrained model is fine-tuned for various downstream tasks at the molecular, protein, and complex levels. Remarkably, EPT showcases strong effectiveness and interpretability in screening potential anti-COVID-19 drugs (Fig.\hyperref[fig1]{1d}).

\subsection{Multi-Domain Pretraining Dataset}
\label{sec:setup}

We construct a large-scale pretraining dataset of over 5.89M entries by integrating existing 3D molecules from GEOM~\cite{axelrod2022geom}, PCQM4Mv2~\cite{hu2021ogb}, PDB~\cite{berman2000protein}, and PDBBind~\cite{wang2005pdbbind}. 
For small molecules, we filter conformations with the top-5 Boltzmann weights for each molecule in GEOM~\cite{axelrod2022geom}, yielding 1.89M conformations. These are combined with 3.38M structures from PCQM4Mv2~\cite{hu2021ogb} to form the pretraining dataset. For proteins, we source experimental structures from the Protein Data Bank (PDB~\cite{berman2000protein}), comprising 77,814 unique sequences along with 600k structural entries. Additionally, we incorporate 22,295 binding pockets from PDBBind~\cite{wang2005pdbbind} to enrich the dataset with information on protein-protein and protein-molecule interactions. During pretraining, we deliberately exclude all property labels to prevent data leakage and ensure unbiased performance evaluation in downstream tasks. The details of the pretraining dataset are illustrated in Fig.~\hyperref[fig5]{5}.

\begin{figure}[htbp]
    \centering
    \includegraphics[width=0.9\textwidth]{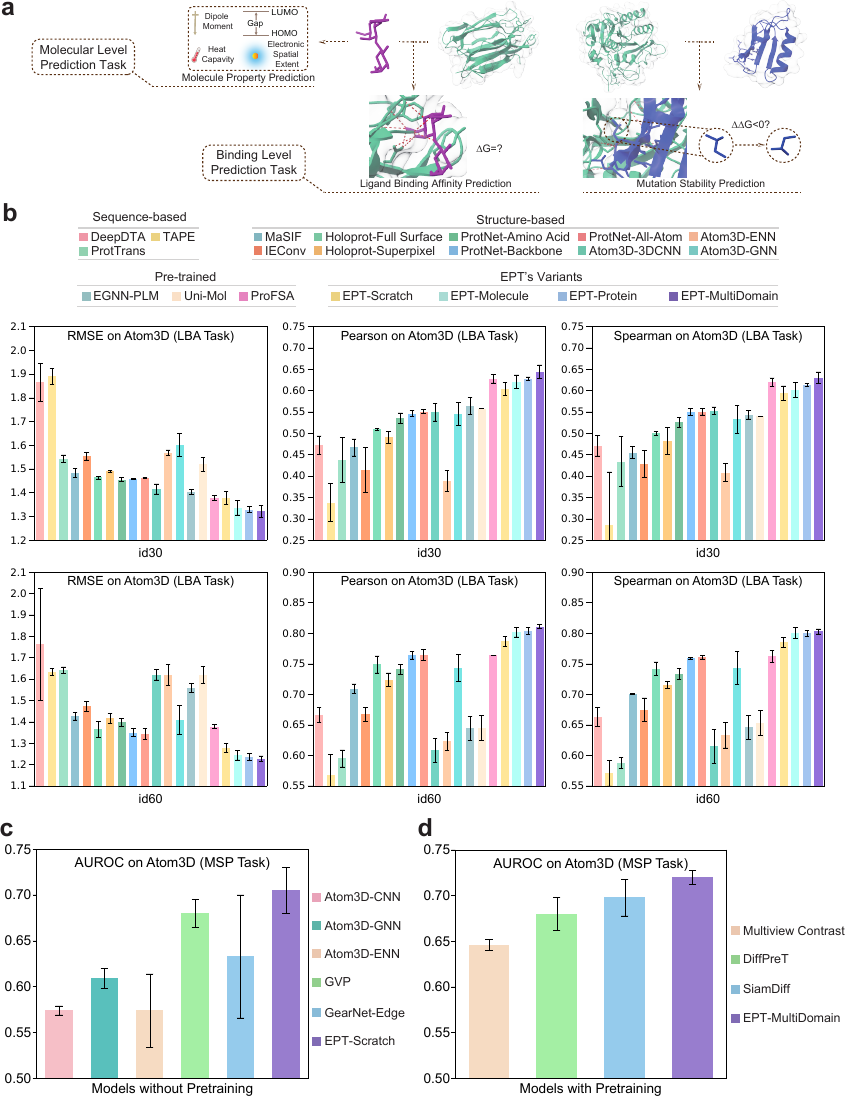}
    \caption{\textbf{\ \textbar \ The performance of EPT on downstream tasks.} \textbf{a}, EPT is evaluated on three downstream tasks: Ligand Binding Affinity Prediction (LBA), Mutation Stability Prediction (MSP) and Molecule Property Prediction (MPP). \textbf{b}, The Root Mean Square Error (RMSE), Pearson and Spearman correlation coefficients on both the id30 and id60 splits for the LBA task. \textbf{c}, The AUROC on Atom3D for the MSP task by the models without pretraining. \textbf{d}, The AUROC on Atom3D for the MSP task by the models with pretraining.}\label{fig2}
\end{figure}

\subsection{Evaluation of model performance on downstream tasks}
\label{sec:downstream}

To evaluate the general performance of EPT, we select typical downstream tasks across two levels (Fig.~\hyperref[fig2]{2a}).  At the binding level, we focus on Ligand Binding Affinity Prediction (LBA) and Mutation Stability Prediction (MSP), while at the molecular level, we choose the Molecule Property Prediction (MPP) task, with details as follows.

\subsubsection{Ligand Binding Affinity Prediction (LBA)}

We address the LBA task using the Atom3D dataset~\cite{townshend2020atom3d}, which involves predicting the binding affinity between a protein pocket and its corresponding ligand. Following the setup in~\cite{townshend2020atom3d}, each sample in the dataset comprises a protein-molecule complex paired with its binding affinity value. The dataset offers two split configurations based on protein sequence similarity thresholds: ID30 (sequence identity capped at 30\%) and ID60 (capped at 60\%). These splits include 3507, 466, and 490 complexes in the training, validation, and test sets, respectively. For evaluation, we employ the Root Mean Square Error (RMSE), Pearson correlation coefficient, and Spearman correlation coefficient as metrics. To ensure the robustness of our results, we conduct experiments using three random seeds and report the mean and standard deviations for each metric.
We compare our method against three categories of prior works: sequence-based methods, including DeepDTA~\cite{ozturk2018deepdta}, B\&B~\cite{bepler2019learning}, TAPE~\cite{rao2019evaluating}, and ProtTrans~\cite{elnaggar2021prottrans}; structure-based models, such as MaSIF~\cite{gainza2020deciphering}, IEConv~\cite{hermosilla2020intrinsic}, Holoprot~\cite{somnath2021multi}, ProtNet~\cite{wang2023learning}, and the three backbone models introduced by Atom3D~\cite{townshend2020atom3d}; and recent pretraining-based methods, including DeepAffinity~\cite{karimi2019deepaffinity}, EGNN-PLM~\cite{wu2022discovering}, Uni-Mol~\cite{zhou2023uni}, and ProFSA~\cite{gao2023self}.

The results in Fig.~\hyperref[fig2]{2b} evaluate our EPT model under four different training conditions: trained from scratch, pretrained on the small molecule subset, pretrained on the protein subset, and pretrained on the entire multi-domain dataset.  We have the following observations: \textbf{1.} Structure-based models generally surpass sequence-based counterparts, underscoring the significance of 3D geometry in capturing interactive information.  \textbf{2.} Pretraining on each individual subset is capable to enhance performance. Remarkably, EPT-Molecule, which is pretrained without exposure to protein or complex structures, still outshines the scratch-trained model, suggesting the presence of cross-domain transferable knowledge.  \textbf{3.} EPT-MultiDomain, benefited from the entire dataset containing diverse domains, outperforms previous methods and achieve state-of-the-art performance on both of the splits. This implies that the breadth of pretraining data correlates positively with the model's performance, and enables a more generalizable understanding of biological interactions.

\subsubsection{Mutation Stability Prediction (MSP)}

Additionally, we evaluate the model on the MSP task to assess its generalization capabilities. This task involves predicting whether a point mutation at the interface of protein complexes improves binding affinity, framed as a binary classification problem. Following~\cite{townshend2020atom3d}, we report AUROC scores on the dataset split with 30\% sequence identity. This split comprises 2864 samples for training, 937 for validation, and 347 for testing.

We compare the scratch-trained EPT with Atom3D-CNN, GNN, and ENN~\cite{townshend2020atom3d}, as well as GVP~\cite{jing2021equivariant} and GearNet-Edge~\cite{zhang2022protein}, as shown in Fig.\hyperref[fig2]{2c}. Additionally, we evaluate the multi-domain pretrained EPT against prior structure-based pretraining strategies using GearNet-Edge~\cite{zhang2022protein}, including Multiview Contrast, DiffPreT, and SiamDiff, as illustrated in Fig.~\hyperref[fig2]{2d}. The results demonstrate that EPT outperforms all baselines, both when trained from scratch and when pretrained on external dataset, highlighting its superior foundational expressiveness and exceptional knowledge transferability in modeling macromolecular systems.

\begin{table*}[h]
\caption{MAEs for the MPP task on the QM9 dataset. The best results are in \textbf{bold} and the second best are \underline{underlined}. The right-most column provides the averaged rank of each method across 12 tasks.}
\resizebox{\linewidth}{!}{
\begin{tabular}{lccccccccccccc}
\midrule
\multirow{2}{*}{Model} & $\mu$$\downarrow$                   & $\alpha$$\downarrow$                      & $\epsilon_{\text{HOMO}}$$\downarrow$         & $\epsilon_{\text{LUMO}}$$\downarrow$        & $\Delta\epsilon$$\downarrow$          & $<R^2>$$\downarrow$                       & ZPVE$\downarrow$                      & $U_0$$\downarrow$                     & $U$$\downarrow$                       & $H$$\downarrow$                       & $G$$\downarrow$                       & $C_v$$\downarrow$          & Avg.$\downarrow$                     \\
                       & \multicolumn{1}{c}{(D)} & \multicolumn{1}{c}{($a_0^3$)} & \multicolumn{1}{c}{(meV)} & \multicolumn{1}{c}{(meV)} & \multicolumn{1}{c}{(meV)} & \multicolumn{1}{c}{($a_0^2$)} & \multicolumn{1}{c}{(meV)} & \multicolumn{1}{c}{(meV)} & \multicolumn{1}{c}{(meV)} & \multicolumn{1}{c}{(meV)} & \multicolumn{1}{c}{(meV)} & \multicolumn{1}{c}{($\frac{cal}{mol K}$)}  & Rank
\\ \midrule

SchNet         & 0.033          & 0.235          & 41.0          & 34.0          & 63.0          & \underline{0.070}    & 1.70          & 14.00         & 19.00         & 14.00         & 14.00         & 0.033           & 11.83         \\
E(n)-GNN       & 0.029          & 0.071          & 29.0          & 25.0          & 48.0          & 0.110          & 1.55          & 11.00         & 12.00         & 12.00         & 12.00         & 0.031          & 11.17         \\
DimeNet++      & 0.030          & 0.043          & 24.6          & 19.5          & 32.6          & 0.330          & 1.21          & 6.32          & 6.28          & 6.53          & 7.56          & 0.023          & 7.17          \\
PaiNN          & 0.012          & 0.045          & 27.6          & 20.4          & 45.7          & \underline{0.070}    & 1.28          & 5.85          & 5.83          & 5.98          & 7.35          & 0.024         & 6.33          \\
TorchMD-Net    & 0.011          & 0.059          & 20.3          & 18.6          & 36.1          & \textbf{0.033} & 1.84          & 6.15          & 6.38          & 6.16          & 7.62          & 0.026        & 7.08          \\
Equiformer     & 0.011          & 0.046          & \textbf{15.0} & 14.0          & 30.0          & 0.251          & 1.26          & 6.59          & 6.74          & 6.63          & 7.63          & 0.023        & 6.00          \\ \midrule
Transformer-M  & 0.037          & \underline{0.041}    & 17.5          & 16.2          & \textbf{27.4} & 0.075          & 1.18          & 9.37          & 9.41          & 9.39          & 9.63          & 0.022         & 6.92          \\
GeoSSL         & 0.015          & 0.046          & 23.5          & 19.5          & 40.2          & 0.122          & 1.31          & 6.92          & 6.99          & 7.09          & 7.65          & 0.024        & 8.42          \\
3D-EMGP        & 0.020          & 0.057          & 21.3          & 18.2          & 37.1          & 0.092          & 1.38          & 8.60          & 8.60          & 8.70          & 9.30          & 0.026        & 8.83          \\
DP-TorchMD-Net & 0.012          & 0.052          & 17.7          & 14.3          & 31.8          & 0.450          & 1.71          & 6.57          & 6.11          & 6.45          & 6.91          & \textbf{0.020}  & 6.67          \\
Frad           & \textbf{0.010} & \textbf{0.037} & 15.3          & \underline{13.7}    & \underline{27.8}    & 0.342          & 1.42          & \textbf{5.33} & \underline{5.62}    & 5.55          & \textbf{6.19} & \textbf{0.020} &  \underline{3.17}    \\ \midrule
EPT            & 0.011          & 0.045          & 16.2          & 14.1          & 29.6          & 0.122          & \underline{1.14}    & 5.53          & 5.70          & \underline{5.52}    & 6.42          & \textbf{0.020}  & 3.33          \\
EPT-10         & \textbf{0.010} & 0.045          & \underline{15.2}    & \textbf{13.6} & 29.0          & 0.152          & \textbf{1.11} & \underline{5.44}    & \textbf{5.54} & \textbf{5.42} & \underline{6.37}    & \textbf{0.020}  & \textbf{2.33} \\ \bottomrule
\end{tabular}
  \label{tab1}
}
\vskip -0.2in
\end{table*}

\begin{figure}[t!]
    \centering
    \includegraphics[width=0.9\textwidth]{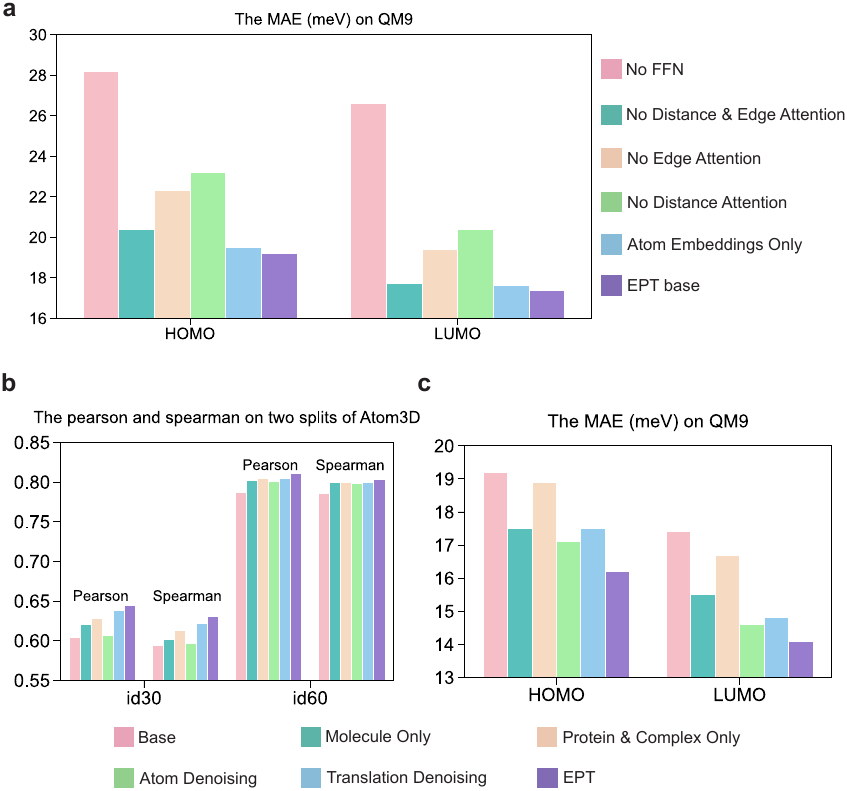}
    \caption{\textbf{\ \textbar \ The validation for EPT's core modules and pretraining strategy.} \textbf{a}, Mean Absolute Errors (MAE) of the HOMO and LUMO predictions on QM9, under different EPT variants.  \textbf{b}, Pearson and Spearman correlation coefficients on id30 and id60 splits for the LBA task, under different pretraining strategies.     
     \textbf{c}, MAE of the HOMO and LUMO predictions on QM9, under different pretraining strategies.
     }\label{fig3}
\end{figure}

\subsubsection{Molecule Property Prediction (MPP)}

For the MPP task, We select QM9~\cite{ramakrishnan2014quantum} to evaluate the performance of EPT on small molecules. In detail, QM9 serves as a quantum chemistry benchmark that offers 12 chemical properties for each 3D molecule composed of C, H, O, N, and F elements. Following~\cite{tholke2022torchmd}, we randomly select 10,000 and 10,831 structures for validation and testing, and the remaining 110,000 structures are used to fine-tune the model. 

To benchmark EPT, we compare it against a variety of 3D geometric models and pretraining approaches designed for small molecules. For geometric GNNs, the evaluation includes SchNet~\cite{schutt2018schnet}, E(n)-GNN~\cite{satorras2021n}, DimeNet++\cite{gasteiger2020fast}, PaiNN\cite{schutt2021equivariant}, and Transformer-based architectures such as TorchMD-Net~\cite{tholke2022torchmd} and Equiformer~\cite{liao2022equiformer}, which leverage vector or higher-order features. Pretraining comparisons are drawn from~\cite{feng2023fractional}, including GeoSSL~\cite{liu2022molecular} and 3D-EMGP~\cite{jiao2023energy}, which apply denoising techniques to PaiNN and E(n)-GNN, as well as Transformer-M~\cite{luo2022one}, DP-TorchMD-Net~\cite{zaidi2022pre}, and Frad~\cite{feng2023fractional}, which utilize various denoising strategies with Transformer-based models. Model performance is evaluated using the Mean Absolute Error (MAE) for each property, and the average rank across 12 tasks is computed to provide a concise summary of the results.

In Table~\ref{tab1}, our EPT outperforms or matches the performance of existing denoising-based methods, underscoring the effectiveness of multi-domain block-level pretraining. Additionally, we evaluate an augmented model, \textbf{EPT-10}, which incorporates 10 layers as opposed to the original 6-layer configuration. The improved results, presented in the last row of Table~\ref{tab1}, demonstrating that performance improves as the model complexity is scaled up.

\subsubsection{Analyses of the core components in EPT}
\label{sec:abl}

We provide a series of ablation studies on the MPP task (QM9 dataset) to elucidate the contribution of each component to the performance of our backbone model, as detailed in Fig.~\hyperref[fig3]{3a}. Specifically, we explore the following aspects. \textbf{1.} We first substitute the input embedding delineated in Eq.~(\ref{eq:emb}) with a straightforward atom-level embedding, denoted as $\vf_i=f_a(a_i)$. The findings suggest that enriching atom features with block-level information slightly improves model performance. \textbf{2.} In Eq.~(\ref{eq:attn}), we integrate the distance matrix $\mathbf{D}$ and edge features $\mathbf{R}$ into the attention mechanism. Eliminating either or both of these elements leads to a decline in performance, thereby underscoring their collective significance in effectively capturing the varying interatomic relations. \textbf{3.} In Eq.~(\ref{eq:ffn1}-\ref{eq:ffn3}), we employ the FFN layer to amalgamate scalar and vector features. The resultant sharp drop after removing the FFN layer underscores the critical role of feature fusion in our model.

In addition, we evaluate the influence of pretraining datasets and denoising strategies on the LBA and MPP tasks in Fig.~\hyperref[fig3]{3b}. As an extension of Fig.~\hyperref[fig2]{2b}, we observe a consistent trend where pretraining on one domain confers benefits to downstream tasks in another domain. Specifically, the model pretrained on small molecules demonstrates enhanced performance on the LBA task, while the model pretrained on proteins exhibits improved results on the QM9 benchmark. In general, the model pretrained on the multi-domain dataset shows superior performance across all evaluated downstream tasks.
We further compare the three kinds of denoising strategies introduced in~\textsection\ref{sec:dn}. The $\gL_{\text{atom}}$ strategy focuses on atom-level denoising, showing superior results for small molecular structures. However, its benefits are less pronounced when applied to large, complex systems such as the protein-molecule complexes in LBA. On the contrary, $\gL_{\text{block-T}}$ adopts a more macroscopic approach by only considering the translations of each block's center of mass. This coarse-grained strategy improves performance for larger systems but tends to struggle in smaller molecules. Finally, our model, which accounts for both block-level rotations and translations, provides a comprehensive supervision for predicted forces, resulting in the optimal performance across all strategies.

\begin{figure}[htbp]
    \centering
    \includegraphics[width=0.9\textwidth]{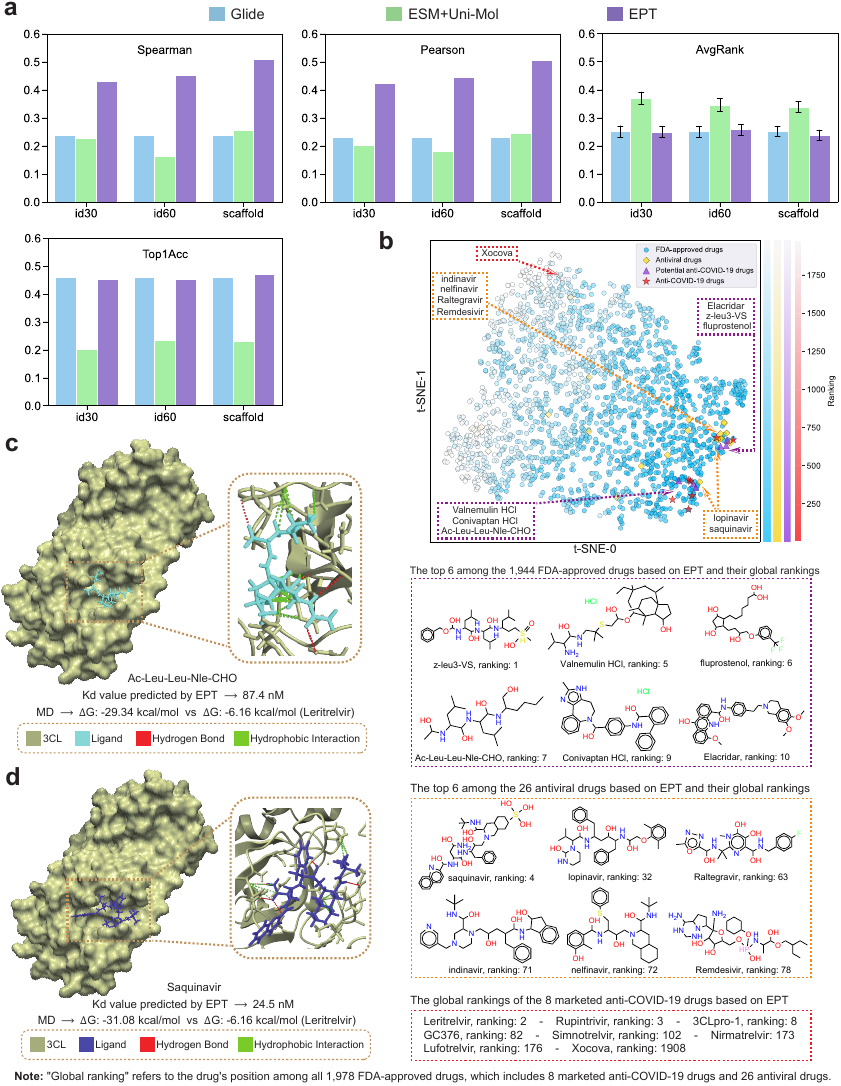}
    \caption{\textbf{\ \textbar \ The application of EPT in screening anti-COVID-19 drugs.} \textbf{a}, Benchmark experiments on the PDBBind dataset: predicting the binding affinity (evaluated by the Pearson and Spearman correlation coefficients), and identifying the positive candidate bound to the target pocket (evaluated by the ranking metrics AvgRank and Top1Acc). \textbf{b}, Visualization of EPT embeddings, rankings of 1,978 FDA-approved drugs based on EPT, and the presentation of focus molecules. \textbf{c}, \textbf{d}, A comprehensive analysis of the two hits screened based on EPT is conducted through molecular docking and MD simulations. Leritrelvir is a positive reference.}\label{fig4}
\end{figure}

\subsection{EPT-accelerated discovery of anti-COVID-19 candidates}

We conduct virtual screening experiments to find potential anti-COVID-19 small-molecule drugs. To this end, we design benchmark experiments using the PDBBind dataset, which is partitioned into three subsets based on different strategies: id30, id60, and scaffold splits. For each partition, we define two evaluation tasks, namely the \textit{affinity prediction} task and the \textit{ranking} task.
The affinity prediction task involves predicting the binding affinity of all ligand-pocket complexes and measuring the Spearman and Pearson correlations between the model's predictions and the ground-truth affinity labels. The ranking task aims to determine whether a given ligand candidate can bind to the target pocket. To achieve this, we randomly sample 10 additional ligands as negative examples for each pocket and train a classification head to distinguish between positive and negative ligands. The classifier outputs are then used to rank the 11 samples (1 positive and 10 negatives). The model’s performance is evaluated using two metrics: the average rank of the positive sample (AvgRank) and the proportion of cases where the positive sample is ranked as the top candidate (Top-1 Accuracy).

We adopt the pretrained EPT and finetune it on the tranining complexes. 
To benchmark EPT, we compare its performance against two typical baselines:  the classical method Glide~\cite{friesner2004glide} based on energy computation, and the dual-tower model ESM+Uni-Mol which applies the pretrained protein model ESM~\cite{lin2022language} for pocket modeling and Uni-Mol~\cite{zhou2023uni} for representing small molecules. As shown in Fig.~\hyperref[fig4]{4a}, our model achieves significantly better performance than both Glide and the dual-tower models in the affinity prediction tasks. For ranking tasks, EPT outperforms the dual-tower models and performs on par with Glide. These results underscore the effectiveness of our unified representation model in capturing the complex interactions between ligands and their target protein pockets.

Based on these findings, we apply our finetuned model to rank 1,978 FDA-approved drugs (including 26 antiviral drugs and 8 marketed anti-COVID-19 drugs) in order to evaluate their binding potential to 3CL protease, a key target in SARS-CoV-2 replication. The rankings of 8 anti-COVID-19 drugs are shown in Fig.~\hyperref[fig4]{4b}, where we observe that nearly all drugs are ranked within the top 200. We visualize the representations of all ligand-pocket complexes using t-SNE in Fig.~\hyperref[fig4]{4b}. The results reveal that, with the exception of the outlier Xocova, anti-COVID-19 drugs generally form a concentrated cluster surrounded by other drugs with high EPT ranking scores.
Additionally, we select the top 6 antiviral drugs based on their rankings and further identify the top 6 non-antiviral, non-anti-COVID-19 drugs as extra potential anti-COVID-19 candidates. These 12 top-ranking candidates are located closely with 7 marketed anti-COVID-19 drugs in the t-SNE embeddings shown in Fig.~\hyperref[fig4]{4b}, indicating their potential high affinity for the 3CL protease. To quantitatively assess the binding potential of the 12 selected candidates, we perform molecular docking and Molecular Dynamics (MD) simulations, identifying two hits, Ac-Leu-Leu-Nle-CHO and Saquinavir, with EPT-predicted dissociation constants ($K_d$) of 87.4\,\text{nM} and 24.5\,\text{nM}, respectively. As exhibited in Fig.~\hyperref[fig4]{4c} and Fig.~\hyperref[fig4]{4d}, on one hand, the docking results indicate that these two hits indeed have some interactions such as hydrogen bonds and hydrophobic interactions with 3CL protease that enhance binding affinities. On the other hand, compared to the positive reference Leritrelvir with $\Delta G~\text{of}~-6.16 \,\text{kcal/mol}$, $\Delta G$ of these two hits are both $\sim-30 \,\text{kcal/mol}$, further suggesting their potential as clinical drugs. Additional examples can be found in Supplementary Fig.~\hyperref[fig:mem]{1}. Of the 12 selected candidates, 9 demonstrate binding potential to the target with $\Delta G<0$, and 7 deliver lower $\Delta G$ than Leritrelvir, the top-ranking anti-COVID-19 drug in Fig.\hyperref[fig4]{4b}. These results not only emphasize the predictive power of EPT but also highlight its potential to accelerate the identification of promising drug candidates.

\section{Discussion}
\label{sec:con}


In this work, we propose EPT, an equivariant transformer-based model pretrained on multi-domain 3D molecular structures. We unify the representation of molecules from different domains by introducing the concept of blocks, enabling the capture of hierarchical information from atomic-level details in small molecules to residue-level features in proteins. Leveraging this, we design a block-level denoising pretraining strategy, where the model learns to recover from geometric perturbations applied to these blocks. To effectively handle this task, we utilize an equivariant Transformer architecture that preserves the intrinsic symmetries of physical laws, ensuring that the model captures the complex geometric relationships while being scalable for large, diverse datasets.

The effectiveness of EPT is demonstrated through its superior performance on a variety of benchmark tasks, including Ligand Binding Affinity (LBA) prediction, Mutation Stability Prediction (MSP) and Molecular Property Prediction (MPP). EPT outperforms existing domain-specific methods in these tasks, highlighting its ability to generalize across different molecular domains. Furthermore, when applied to predict the binding potential of FDA-approved drugs to the 3CL protease, a critical target in the fight against SARS-CoV-2, EPT identified known anti-COVID-19 drugs as top candidates and facilitated the identification of promising new drug candidates, showcasing its real-world applicability.

The development of EPT opens exciting possibilities for future research and applications in molecular science. Future work could involve extending the model to other molecular systems like crystals and catalysts and further enhancing its denoising techniques to handle even more complex molecular interactions. The success of EPT also paves the way for future foundation models in molecular science, where unified, scalable models can drive breakthroughs across a broad range of scientific domains, from drug design to protein engineering. 

\section*{Methods}

\begin{figure}[htbp]
    \centering
    \includegraphics[width=0.9\textwidth]{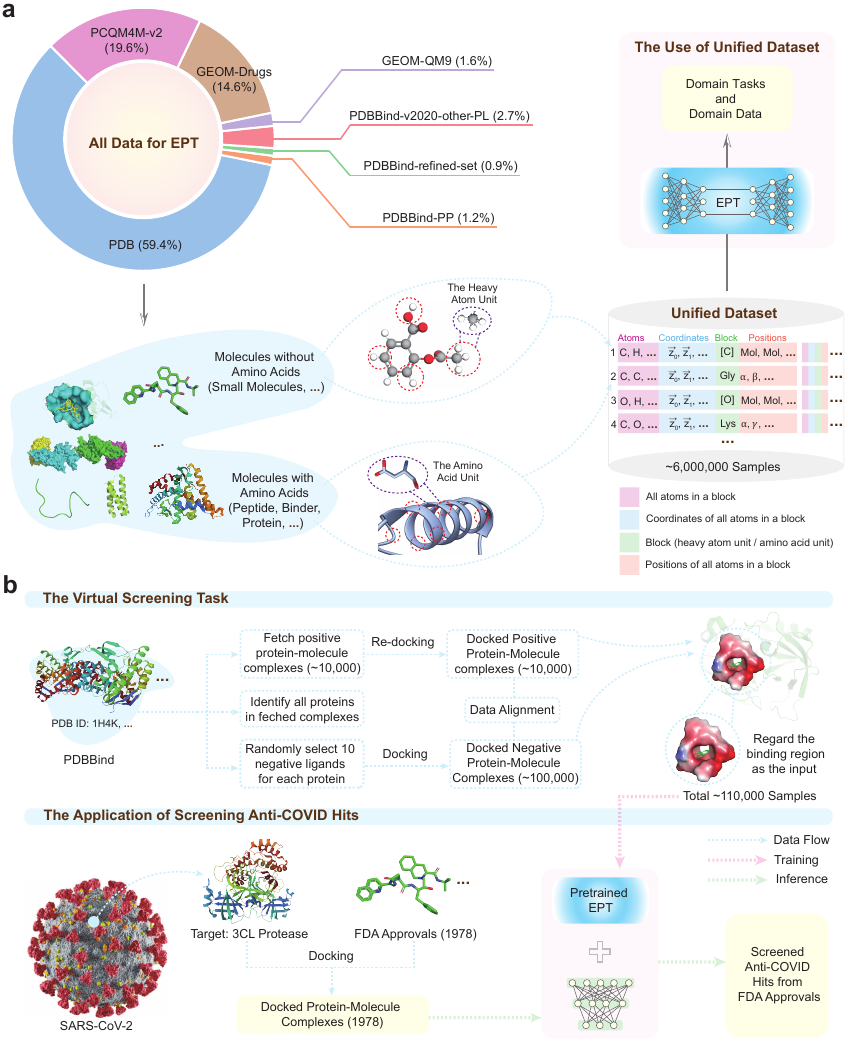}
    \caption{\textbf{\ \textbar \ Data description and preprocessing.} \textbf{a}, The pretraining dataset combines small molecule conformations, protein structures as well as protein-protein and protein-molecule complexes. Each entry is represented as a list of blocks, with each block characterized by four features: the atom types, the block type, the ordered position indexes, and the atom coordinates.
    \textbf{b}, We illustrate the overall pipeline for evaluating the virtual screening task. Fine-tuning of the pretrained EPT model is performed using a curated dataset of protein-small molecule complexes collected from PDBBind and re-docked to generate positive and negative samples. The finetuned model is then applied to rank FDA-approved drugs based on their predicted binding probability from the docked complex with the 3CL protease.}\label{fig5}
\end{figure}

\subsection*{Datasets}
\label{sec:datasets}


We curate a comprehensive structural dataset comprising 5.89 million samples and 261 million blocks, spanning small molecules, proteins, and complexes (Fig.\hyperref[fig5]{5a}). Among these, small molecule datasets, including GEOM~\cite{axelrod2022geom} and PCQM4Mv2~\cite{hu2021ogb}, contribute 35.8\% of the total blocks. For protein structures, experimental data are collected from the Protein Data Bank (PDB~\cite{berman2000protein}), which encompasses 600,000 structural entries and accounts for 59.4\% of the blocks. The remaining 4.8\% of blocks are derived from PDBBind~\cite{wang2005pdbbind}, including protein-protein complexes (PP), protein-molecule complexes from the refined set, and other protein-molecule complexes collected before version 2020, denoted as v2020-other-PL. Each sample is stored as a sequence of blocks, with each block defined by four attributes: the types of atoms within the block, the type of the block itself, the position index of each atom, and the 3D coordinate matrix of the block.


We employ virtual screening as a typical application of the pretrained EPT model (Fig.~\hyperref[fig5]{5b}). To fine-tune EPT, we first prepare the Docked-PDBBind dataset. Specifically, we obtain 10,000 protein-small molecule complexes from PDBBind and re-dock them using the Glide software~\cite{friesner2004glide} to generate 10,000 positive samples. Additionally, for each complex, we randomly selected 10 non-binding small molecules and docked them with the protein using Glide, yielding 100,000 negative samples. The re-docking process ensures consistency with the inference phase, where only individual protein and small molecule structures are available, rather than pre-formed complexes. This alignment is crucial to maintain distributional similarity between the training data and the docking results used during inference. For inference, we dock the target 3CL protease with 1,978 FDA-approved drugs and rank them based on the predicted binding probabilities from the fine-tuned model.

\subsection*{Model Architecture}

We represent a molecule with $N$ atoms as a fully-connected graph $\gG$, where atoms are depicted as nodes and their interactions are depicted as edges $\gE$. To capture the high-level structure within molecules, atoms are grouped into $M$ predefined blocks to enrich the node features~\cite{kong2023generalist}. Let $m_i$ denote the index of the block containing atom $i$, the feature set for an atom is extended to $(a_i, b_{m_i}, p_i, \vec{\vz}_i)$, where $a_i \in \gA$ specifying the atom type, $b_{m_i}$ indicates the block type of $m_i$, $p_i$ denotes the atom's positional index within its block, and $\vec{\vz}_i \in \sR^3$ representing the atom's 3D coordinate.  The positional index within an amino acid is determined using Greek letters, ordered based on their distance from the C$_\alpha$ atom. For small molecules, the positional index is represented by a designated token, [Mol].
Interactions between atoms are categorized into three distinct edge types to reflect both intra-block and inter-block relationships. Mathematically,
\begin{equation}e_{ij}=
\begin{cases}
0, &m_i=m_j,\\
1, &m_i\neq m_j, d(m_i, m_j)\leq \delta_{\text{topo}},\\
2, &m_i\neq m_j, \delta_{\text{topo}} < d(m_i, m_j)\leq \delta_{\text{max}},\\
\end{cases}
\end{equation}
where $\delta_{\text{topo}}$ and $\delta_{\text{max}}$ are predefined thresholds that represent topological and maximum allowable distances, respectively. The function $d(m_i, m_j) = \min_{m_p=m_i,m_q=m_j}\|\vec{z}_p - \vec{z}_q\|_2$ calculates the minimum Euclidean distance between any two atoms belonging to blocks $m_i$ and $m_j$. 

In the pursuit of efficiently capturing the nuanced interactions of atoms within molecules, we present the Equivariant Full-Atom Transformer. It utilizes the Transformer-based backbone~\cite{vaswani2017attention} to model the complex interactions among atoms, while updating the scalar and vector features to capture the rich geometric information inherent in molecular structures. 

Our model first acquires the input features from the Graph Embedding layer, and iteratively updates the features at each layer $l$. Let $\mH^{(l)}=[\vh_1^{(l)}, \vh_2^{(l)}, \cdots, \vh_N^{(l)}]\in\sR^{h\times N}$ denote the scalar, and $\vec{\mV}^{(l)}=[\vec{\vv}_1^{(l)}, \vec{\vv}_2^{(l)}, \cdots, \vec{\vv}_N^{(l)}]\in\sR^{3\times h\times N}$ denote the vector. The model is constructed in this way:
\begin{align}
    [\mH^{(0)}, \vec{\mV}^{(0)}] &= \text{Embedding}(\mA, \mB, \mP, \vec{\mZ}), \\
    [\mH^{(l-0.5)}, \vec{\mV}^{(l-0.5)}] &= [\mH^{(l-1)}, \vec{\mV}^{(l-1)}] + \text{Self-Attn}(\text{LN}(\mH^{(l-1)}), \vec{\mV}^{(l-1)}), \\
    [\mH^{(l)}, \vec{\mV}^{(l)}] &= [\mH^{(l-0.5)}, \vec{\mV}^{(l-0.5)}] + \text{FFN}(\text{LN}(\mH^{(l-0.5)}), \vec{\mV}^{(l-0.5)}).  
\end{align}

After the Embedding layer, Self-attention (Self-Attn) and Feed-Forward Networks (FFN) are applied alternately, with pre-layer normalization (LN) and residual connections preceding each operation. We modify the Self-Attn and FFN layers to be E(3)-equivariant, preserving the geometrical symmetry of molecular structures. These layers are detailed as follows.  For conciseness, we omit the layer subscribe $l$ unless otherwise specified.

For the graph embedding layer, the input features are obtained as:
\begin{align}
\label{eq:emb}
    \vf_i &=f_b(b_{m_i}) + f_a(a_i) + f_p(p_i), \\
    \ve'_{ij} &= [\vf_i, \vf_j, \ve_{ij}, \text{RBF}(\|\vec{\vz}_i-\vec{\vz}_j\|_2)],\\
    \vh_i^{(0)} &= \varphi_h(\vf_i, \sum_{j\in\gN(i)}\varphi_s(\ve'_{ij})\cdot\vf_j), \\
    \label{eq:v_init}
    \vv_i^{(0)} &= \sum_{j\in\gN(i)}\varphi_v(\ve'_{ij})\cdot(\vec{\vz}_i-\vec{\vz}_j),
\end{align}
where $f_b, f_a, f_p$ separately embed the block types, atom types and atom orders, $\text{RBF}(\cdot)$ denote the radial basis functions, and $\varphi_h, \varphi_s, \varphi_v$ are MLPs to aggregate neighbor information to enrich the $0$-th layer features. It is necessary to initialize $\vec{\mV}^{(0)}$ with SE(3)-equivariant non-zero values via Eq.~(\ref{eq:v_init}), otherwise the vector features will remain zeros in subsequent layers.

The self-attention layer plays a crucial role in modeling interatomic interactions. For each layer, query $\mQ_s$, key $\mK_s$, and value $\vec{\mV}_s$ matrices of the $s$-th head are computed as
\begin{align}
\label{eq:attn_begin}
    \mQ_s=\mH\mW_s^Q, \mK_s=\mH\mW_s^K, \vec{\mV}_s=[\mH\mW_s^{Vh}, \vec{\mV}\mW_s^{Vv}], 
\end{align}
where $h_s$ is the dimension of each head, $\mW_s^Q, \mW_s^K\in\sR^{h\times 4h_s}, \mW_s^{Vh}, \mW_s^{Vv}\in\sR^{h\times h_s}$ are trainable parameters that map the features to the appropriate query, key, and value spaces. And the attention mechanism is given by
\begin{align}
\label{eq:attn}
    \mH_s,\vec{\mV}_s=\text{Softmax}\Big(\frac{\mathbf{Q_s^\top K_s}}{2\sqrt{h_s}} - \mD  + \mR\Big)\vec{\mV}_s,
\end{align}
where $\mD=\{d_{ij}\}_{i,j=1}^N=\{\|\vec{\vz}_i-\vec{\vz}_j\|_2\}_{i,j=1}^N$ is the distance matrix, and $\mR=\{r_{ij}\}_{i,j=1}^N$ encodes the edge interactions and geometric relations as
\begin{equation}r_{ij}=
\begin{cases}
\varphi_r(\ve_{ij}, \text{RBF}(\|\vec{\vz}_i-\vec{\vz}_j\|_2)), &(i,j)\in\gE,\\
0, &(i,j)\notin\gE.
\end{cases}
\end{equation}
Here $\varphi_r$ is an MLP. The outputs of the self-attention layer combines the contributions of all heads:
\begin{align}
\label{eq:attn_end}
    \Delta\mH=\sum_s \mH_s\mW_s^{Oh}, \Delta\vec{\mV}=\sum_s \vec{\mV}_s\mW_s^{Ov},
\end{align}
where $\mW_s^{Oh}, \mW_s^{Ov}\in\sR^{h_s\times h}$ are head-specific trainable parameters.

Building upon the Geometric Vector Perceptron (GVP~\cite{jing2021equivariant}) concept, the equivariant feed-forward layer is where the scalar and vector features are fused and updated simultaneously:
\begin{align}
\label{eq:ffn1}
    \vec{\mV}_1, \vec{\mV}_2 &=\vec{\mV}\mW_1, \vec{\mV}\mW_2,\\
    \Delta\mH, \mU &= \varphi_{\text{FFN}}(\mH, \|\vec{\mV}_1\|_2),\\ \label{eq:ffn3}
    \Delta\vec{\mV} &= \text{LN}(\mU)\odot\vec{\mV}_2,
\end{align}
where $\mW_1, \mW_2\in\sR^{h\times h}$ are learnable linear projectors, $\varphi_{\text{FFN}}$ is an MLP that integrates the scalar features with the magnitude of the vector features, and $\odot$ denotes element-wise multiplication. The intermediate matrix $\mU$ is layer-normalized to conserve the scale of the updated vectors.

\subsection{Block-level denoised pretraining}
\label{sec:dn}

Based on the backbone model above, an additional force prediction head is required for denoised pretraining. In practice, we apply an additional FFN-like layer over layer-L to fuse the output scalars and vectors as
\begin{align}
    \mH_{\text{out}}, \vec{\mV}_{\text{out}} &= \mH^{(L)}, \vec{\mV}^{(L)} / \|\vec{\mV}^{(L)}\|_2,\\
    \vec{\mF}'&=\varphi_{\text{out}}(\mH_{\text{out}}, \|\vec{\mV}_{\text{out}}\mW_1'\|_2) \odot \vec{\mV}_{\text{out}}\mW_2'.
\end{align}

To encode hierarchical molecular information, we introduce a block-level denoising pretraining strategy that integrates both translational and rotational perturbations applied to molecular blocks. This approach considers blocks as rigid bodies, where all atoms within a block are perturbed together, effectively capturing both intra- and inter-block geometric relationships.

Perturbations are applied by introducing translational and rotational noise to each block, modeled as:
\begin{align}
    \vec{\mZ}' = C\Big(g_b(\vec{\mZ}_b + \sigma_t\vepsilon_{\mZ_b}) + \mQ_b\vec{\mZ}_r\Big),
\end{align}
where $\vepsilon_{\mZ_b}\sim\gN(0,\mI_{3M})$ represents translational noise applied at the block level, $g_b(\cdot)$ maps block coordinates to atoms, $\mQ_b$ is the rotation matrix derived from rotational noise $\vomega_b$ sampled from the isotropic Gaussian distribution $\mathcal{IG}_{SO(3)}(\sigma_r)$~\citep{leach2022denoising}, and $\vec{\mZ}_r=\vec{\mZ} - g_b(\vec{\mZ}_b)$ represents the relative atom positions to block centers. The centering operator $C(\cdot)$ ensures translational neutrality.

The pretraining objective incorporates two components. First, a translation objective minimizes the discrepancy between predicted resultant forces and translational noises averaged over blocks:
\begin{align}
\label{eq:prac_blockt_loss}
    \gL_{\text{block-T}} &= \E_{\vepsilon_{\mZ_b}\sim\gN(0,\mI_{3M})}\Big[\|\mu_b(\vec{\mF}')-\frac{\mu_b(\vec{\mZ}') - \vec{\mZ}_b}{\sigma_t^2}\|_2^2\Big].
\end{align}
where $\mu_b(\cdot)$ computes block-level averages. Second, a rotation objective aligns angular accelerations derived from predicted torques with gradients of the rotational noise distribution.
\begin{align}
    \gL_{\text{block-R}} = \E_{\vomega\sim\mathcal{IG}_{SO(3)}(\sigma_r)}\Big[\|\vec{\valpha}_b - \nabla_{\vomega}p(\vomega)\|_2^2\Big].
\end{align}
where $\vec{\valpha}_b=\mI_b^{-1} \vec{\mM}'_b$ is the predicted angular acceleration, $\vec{\mM}'_b = \sum_j \big(\vec{\vz}_j - g_b(\vec{\mZ}_{b})\big)\times\vec{\vf}'_j$ represents the torque, and $\mI_b$ is the inertia matrix of each block.
The combined block-level denoising loss integrates these two objectives:
\begin{align}
    \gL_{\text{block-C}} = \gL_{\text{block-T}} + \gL_{\text{block-R}}.
\end{align}

\subsection*{Training Regimen}

We first pretrain the backbone model on the collected multi-domain dataset based on the objective $\gL_{\text{block-C}}$, leveraging 8 NVIDIA Tesla A800 GPUs. The pretraining process spans 50 epochs, incorporating translational and rotational noise scales of $\sigma_t=0.04$ and $\sigma_r=0.1$, respectively. A dynamic batching strategy is applied, capping each mini-batch at a maximum of 10,000 atoms per GPU. After pretraining, the force prediction head is omitted, and the backbone model is further finetuned with task-specific heads. 

For MPP, we utilize the noisy node technique~\cite{godwin2021simple} by adding $\gL_\text{block-C}$ as an auxiliary training objective, and the entire loss for finetuning on QM9 can be formulated as $\gL=\gL_\text{MAE}+\lambda \gL_\text{block-C}$, where $\lambda=0.1$ balances the weight of each term.

For LBA, we consider three types of output heads based on an MLP $\varphi_E$ as follows:
\begin{align}
    \varphi_\text{atom}(\mH^{(l)}) &= \sum_i \varphi_E(\vh_i^{(l)}),\\
    \varphi_\text{block}(\mH^{(l)}) &= \sum_{m_i} \varphi_E\Big(\sum_{m_j=m_i}\vh_j^{(l)})\Big),\\
    \varphi_\text{graph}(\mH^{(l)}) &= \varphi_E(\sum_i\vh_i^{(l)}).
\end{align}

For MSP, We use the split by sequence identity over 30\% provided by Atom3D~\citep{townshend2020atom3d}, and extract all residues within 6\AA~distance to the mutation point as the local view for input, where the distance between two residues is measured by the minimum distance between atom pairs. 

\subsection*{Evaluation Metrics}

We use Mean Absolute Error (MAE) on the MPP task by measuring the average deviation between predicted properties and ground-truth labels. For the LBA task, we evaluate Pearson Correlation Coefficient (Pearson or PCC), Spearman Correlation Coefficient (Spearman or SPCC), and Root Mean Square Error (RMSE). PCC and SPCC assess the linear and rank-based correlations between predicted and ground-truth affinities, while RMSE quantifies the prediction errors. For the MSP task, we compute the Area Under the Receiver Operating Characteristic curve (AUROC) to measure binary classification performance.

\section*{Author Contributions}

R.J., X.K. and Z.Y. collected and preprocessed the datasets. R.J. and X.K. designed the backbone model and conducted the pretraining and downstream experiments. Z.Y., L.Z., F.R. and W.T. conducted the virtual screening experiment. R.J., X.K., L.Z., Z.Y. and W.H. wrote the manuscript. The study was supervised by W.H. and Y.L..

\putbib[ref]

\label{LastMainPage}
\end{bibunit}



\begin{thebibliography}{10}

\bibitem{axelrod2022geom}
Simon Axelrod and Rafael Gomez-Bombarelli.
\newblock Geom, energy-annotated molecular conformations for property prediction and molecular generation.
\newblock {\em Scientific Data}, 9(1):185, 2022.

\bibitem{bepler2019learning}
Tristan Bepler and Bonnie Berger.
\newblock Learning protein sequence embeddings using information from structure.
\newblock {\em arXiv preprint arXiv:1902.08661}, 2019.

\bibitem{berman2000protein}
Helen~M Berman, John Westbrook, Zukang Feng, Gary Gilliland, Talapady~N Bhat, Helge Weissig, Ilya~N Shindyalov, and Philip~E Bourne.
\newblock The protein data bank.
\newblock {\em Nucleic acids research}, 28(1):235--242, 2000.

\bibitem{blanco2023role}
Alexandre Blanco-Gonzalez, Alfonso Cabezon, Alejandro Seco-Gonzalez, Daniel Conde-Torres, Paula Antelo-Riveiro, Angel Pineiro, and Rebeca Garcia-Fandino.
\newblock The role of ai in drug discovery: challenges, opportunities, and strategies.
\newblock {\em Pharmaceuticals}, 16(6):891, 2023.

\bibitem{brown2020language}
Tom Brown, Benjamin Mann, Nick Ryder, Melanie Subbiah, Jared~D Kaplan, Prafulla Dhariwal, Arvind Neelakantan, Pranav Shyam, Girish Sastry, Amanda Askell, et~al.
\newblock Language models are few-shot learners.
\newblock {\em Advances in neural information processing systems}, 33:1877--1901, 2020.

\bibitem{elnaggar2021prottrans}
Ahmed Elnaggar, Michael Heinzinger, Christian Dallago, Ghalia Rehawi, Yu~Wang, Llion Jones, Tom Gibbs, Tamas Feher, Christoph Angerer, Martin Steinegger, et~al.
\newblock Prottrans: Toward understanding the language of life through self-supervised learning.
\newblock {\em IEEE transactions on pattern analysis and machine intelligence}, 44(10):7112--7127, 2021.

\bibitem{eslami2022artificial}
Mohammed Eslami, Aaron Adler, Rajmonda~S Caceres, Joshua~G Dunn, Nancy Kelley-Loughnane, Vanessa~A Varaljay, and Hector~Garcia Martin.
\newblock Artificial intelligence for synthetic biology.
\newblock {\em Communications of the ACM}, 65(5):88--97, 2022.

\bibitem{feng2023protein}
Shikun Feng, Minghao Li, Yinjun Jia, Weiying Ma, and Yanyan Lan.
\newblock Protein-ligand binding representation learning from fine-grained interactions.
\newblock {\em arXiv preprint arXiv:2311.16160}, 2023.

\bibitem{feng2023fractional}
Shikun Feng, Yuyan Ni, Yanyan Lan, Zhi-Ming Ma, and Wei-Ying Ma.
\newblock Fractional denoising for 3d molecular pre-training.
\newblock In {\em International Conference on Machine Learning}, pages 9938--9961. PMLR, 2023.

\bibitem{friesner2004glide}
Richard~A Friesner, Jay~L Banks, Robert~B Murphy, Thomas~A Halgren, Jasna~J Klicic, Daniel~T Mainz, Matthew~P Repasky, Eric~H Knoll, Mee Shelley, Jason~K Perry, et~al.
\newblock Glide: a new approach for rapid, accurate docking and scoring. 1. method and assessment of docking accuracy.
\newblock {\em Journal of medicinal chemistry}, 47(7):1739--1749, 2004.

\bibitem{gainza2020deciphering}
Pablo Gainza, Freyr Sverrisson, Frederico Monti, Emanuele Rodola, D~Boscaini, MM~Bronstein, and BE~Correia.
\newblock Deciphering interaction fingerprints from protein molecular surfaces using geometric deep learning.
\newblock {\em Nature Methods}, 17(2):184--192, 2020.

\bibitem{gao2023self}
Bowen Gao, Yinjun Jia, Yuanle Mo, Yuyan Ni, Weiying Ma, Zhiming Ma, and Yanyan Lan.
\newblock Self-supervised pocket pretraining via protein fragment-surroundings alignment.
\newblock {\em arXiv preprint arXiv:2310.07229}, 2023.

\bibitem{gasteiger2020fast}
Johannes Gasteiger, Shankari Giri, Johannes~T Margraf, and Stephan G{\"u}nnemann.
\newblock Fast and uncertainty-aware directional message passing for non-equilibrium molecules.
\newblock {\em arXiv preprint arXiv:2011.14115}, 2020.

\bibitem{godwin2021simple}
Jonathan Godwin, Michael Schaarschmidt, Alexander~L Gaunt, Alvaro Sanchez-Gonzalez, Yulia Rubanova, Petar Veli{\v{c}}kovi{\'c}, James Kirkpatrick, and Peter Battaglia.
\newblock Simple gnn regularisation for 3d molecular property prediction and beyond.
\newblock In {\em International Conference on Learning Representations}, 2021.

\bibitem{hermosilla2020intrinsic}
Pedro Hermosilla, Marco Sch{\"a}fer, Matej Lang, Gloria Fackelmann, Pere-Pau V{\'a}zquez, Barbora Kozlikova, Michael Krone, Tobias Ritschel, and Timo Ropinski.
\newblock Intrinsic-extrinsic convolution and pooling for learning on 3d protein structures.
\newblock In {\em International Conference on Learning Representations}, 2020.

\bibitem{hoogeboom2022equivariant}
Emiel Hoogeboom, V{\i}ctor~Garcia Satorras, Cl{\'e}ment Vignac, and Max Welling.
\newblock Equivariant diffusion for molecule generation in 3d.
\newblock In {\em International conference on machine learning}, pages 8867--8887. PMLR, 2022.

\bibitem{hu2021ogb}
Weihua Hu, Matthias Fey, Hongyu Ren, Maho Nakata, Yuxiao Dong, and Jure Leskovec.
\newblock Ogb-lsc: A large-scale challenge for machine learning on graphs.
\newblock {\em arXiv preprint arXiv:2103.09430}, 2021.

\bibitem{jiao2023energy}
Rui Jiao, Jiaqi Han, Wenbing Huang, Yu~Rong, and Yang Liu.
\newblock Energy-motivated equivariant pretraining for 3d molecular graphs.
\newblock In {\em Proceedings of the AAAI Conference on Artificial Intelligence}, volume~37, pages 8096--8104, 2023.

\bibitem{jin2023unsupervised}
Wengong Jin, Siranush Sarkizova, Xun Chen, Nir Hacohen, and Caroline Uhler.
\newblock Unsupervised protein-ligand binding energy prediction via neural euler's rotation equation.
\newblock {\em arXiv preprint arXiv:2301.10814}, 2023.

\bibitem{jing2021equivariant}
Bowen Jing, Stephan Eismann, Pratham~N Soni, and Ron~O Dror.
\newblock Equivariant graph neural networks for 3d macromolecular structure.
\newblock {\em arXiv preprint arXiv:2106.03843}, 2021.

\bibitem{karimi2019deepaffinity}
Mostafa Karimi, Di~Wu, Zhangyang Wang, and Yang Shen.
\newblock Deepaffinity: interpretable deep learning of compound--protein affinity through unified recurrent and convolutional neural networks.
\newblock {\em Bioinformatics}, 35(18):3329--3338, 2019.

\bibitem{kenton2019bert}
Jacob Devlin Ming-Wei~Chang Kenton and Lee~Kristina Toutanova.
\newblock Bert: Pre-training of deep bidirectional transformers for language understanding.
\newblock In {\em Proceedings of NAACL-HLT}, pages 4171--4186, 2019.

\bibitem{kong2023end}
Xiangzhe Kong, Wenbing Huang, and Yang Liu.
\newblock End-to-end full-atom antibody design.
\newblock {\em arXiv preprint arXiv:2302.00203}, 2023.

\bibitem{kong2023generalist}
Xiangzhe Kong, Wenbing Huang, and Yang Liu.
\newblock Generalist equivariant transformer towards 3d molecular interaction learning.
\newblock {\em arXiv preprint arXiv:2306.01474}, 2023.

\bibitem{leach2022denoising}
Adam Leach, Sebastian~M Schmon, Matteo~T. Degiacomi, and Chris~G. Willcocks.
\newblock Denoising diffusion probabilistic models on {SO}(3) for rotational alignment.
\newblock In {\em ICLR 2022 Workshop on Geometrical and Topological Representation Learning}, 2022.

\bibitem{liao2022equiformer}
Yi-Lun Liao and Tess Smidt.
\newblock Equiformer: Equivariant graph attention transformer for 3d atomistic graphs.
\newblock {\em arXiv preprint arXiv:2206.11990}, 2022.

\bibitem{lin2022language}
Zeming Lin, Halil Akin, Roshan Rao, Brian Hie, Zhongkai Zhu, Wenting Lu, Nikita Smetanin, Allan dos Santos~Costa, Maryam Fazel-Zarandi, Tom Sercu, Sal Candido, et~al.
\newblock Language models of protein sequences at the scale of evolution enable accurate structure prediction.
\newblock {\em bioRxiv}, 2022.

\bibitem{liu2022molecular}
Shengchao Liu, Hongyu Guo, and Jian Tang.
\newblock Molecular geometry pretraining with se (3)-invariant denoising distance matching.
\newblock {\em arXiv preprint arXiv:2206.13602}, 2022.

\bibitem{liu2021pre}
Shengchao Liu, Hanchen Wang, Weiyang Liu, Joan Lasenby, Hongyu Guo, and Jian Tang.
\newblock Pre-training molecular graph representation with 3d geometry.
\newblock In {\em International Conference on Learning Representations}, 2021.

\bibitem{luo2022one}
Shengjie Luo, Tianlang Chen, Yixian Xu, Shuxin Zheng, Tie-Yan Liu, Liwei Wang, and Di~He.
\newblock One transformer can understand both 2d \& 3d molecular data.
\newblock {\em arXiv preprint arXiv:2210.01765}, 2022.

\bibitem{ozturk2018deepdta}
Hakime {\"O}zt{\"u}rk, Arzucan {\"O}zg{\"u}r, and Elif Ozkirimli.
\newblock Deepdta: deep drug--target binding affinity prediction.
\newblock {\em Bioinformatics}, 34(17):i821--i829, 2018.

\bibitem{puny2021frame}
Omri Puny, Matan Atzmon, Edward~J Smith, Ishan Misra, Aditya Grover, Heli Ben-Hamu, and Yaron Lipman.
\newblock Frame averaging for invariant and equivariant network design.
\newblock In {\em International Conference on Learning Representations}, 2021.

\bibitem{pyzer2022accelerating}
Edward~O Pyzer-Knapp, Jed~W Pitera, Peter~WJ Staar, Seiji Takeda, Teodoro Laino, Daniel~P Sanders, James Sexton, John~R Smith, and Alessandro Curioni.
\newblock Accelerating materials discovery using artificial intelligence, high performance computing and robotics.
\newblock {\em npj Computational Materials}, 8(1):84, 2022.

\bibitem{radford2018improving}
Alec Radford, Karthik Narasimhan, Tim Salimans, Ilya Sutskever, et~al.
\newblock Improving language understanding by generative pre-training, 2018.

\bibitem{radford2019language}
Alec Radford, Jeffrey Wu, Rewon Child, David Luan, Dario Amodei, Ilya Sutskever, et~al.
\newblock Language models are unsupervised multitask learners, 2019.

\bibitem{ramakrishnan2014quantum}
Raghunathan Ramakrishnan, Pavlo~O Dral, Matthias Rupp, and O~Anatole Von~Lilienfeld.
\newblock Quantum chemistry structures and properties of 134 kilo molecules.
\newblock {\em Scientific data}, 1(1):1--7, 2014.

\bibitem{rao2019evaluating}
Roshan Rao, Nicholas Bhattacharya, Neil Thomas, Yan Duan, Peter Chen, John Canny, Pieter Abbeel, and Yun Song.
\newblock Evaluating protein transfer learning with tape.
\newblock {\em Advances in neural information processing systems}, 32, 2019.

\bibitem{reed2022generalist}
Scott Reed, Konrad Zolna, Emilio Parisotto, Sergio~G{\'o}mez Colmenarejo, Alexander Novikov, Gabriel Barth-maron, Mai Gim{\'e}nez, Yury Sulsky, Jackie Kay, Jost~Tobias Springenberg, et~al.
\newblock A generalist agent.
\newblock {\em Transactions on Machine Learning Research}, 2022.

\bibitem{satorras2021n}
V{\i}ctor~Garcia Satorras, Emiel Hoogeboom, and Max Welling.
\newblock E (n) equivariant graph neural networks.
\newblock In {\em International conference on machine learning}, pages 9323--9332. PMLR, 2021.

\bibitem{schutt2021equivariant}
Kristof Sch{\"u}tt, Oliver Unke, and Michael Gastegger.
\newblock Equivariant message passing for the prediction of tensorial properties and molecular spectra.
\newblock In {\em International Conference on Machine Learning}, pages 9377--9388. PMLR, 2021.

\bibitem{schutt2018schnet}
Kristof~T Sch{\"u}tt, Huziel~E Sauceda, P-J Kindermans, Alexandre Tkatchenko, and K-R M{\"u}ller.
\newblock Schnet--a deep learning architecture for molecules and materials.
\newblock {\em The Journal of Chemical Physics}, 148(24), 2018.

\bibitem{somnath2021multi}
Vignesh~Ram Somnath, Charlotte Bunne, and Andreas Krause.
\newblock Multi-scale representation learning on proteins.
\newblock {\em Advances in Neural Information Processing Systems}, 34:25244--25255, 2021.

\bibitem{stark20223d}
Hannes St{\"a}rk, Dominique Beaini, Gabriele Corso, Prudencio Tossou, Christian Dallago, Stephan G{\"u}nnemann, and Pietro Li{\`o}.
\newblock 3d infomax improves gnns for molecular property prediction.
\newblock In {\em International Conference on Machine Learning}, pages 20479--20502. PMLR, 2022.

\bibitem{tholke2022torchmd}
Philipp Th{\"o}lke and Gianni De~Fabritiis.
\newblock Torchmd-net: equivariant transformers for neural network based molecular potentials.
\newblock {\em arXiv preprint arXiv:2202.02541}, 2022.

\bibitem{thomas2018tensor}
Nathaniel Thomas, Tess Smidt, Steven Kearnes, Lusann Yang, Li~Li, Kai Kohlhoff, and Patrick Riley.
\newblock Tensor field networks: Rotation-and translation-equivariant neural networks for 3d point clouds.
\newblock {\em arXiv preprint arXiv:1802.08219}, 2018.

\bibitem{townshend2020atom3d}
Raphael~JL Townshend, Martin V{\"o}gele, Patricia Suriana, Alexander Derry, Alexander Powers, Yianni Laloudakis, Sidhika Balachandar, Bowen Jing, Brandon Anderson, Stephan Eismann, et~al.
\newblock Atom3d: Tasks on molecules in three dimensions.
\newblock {\em arXiv preprint arXiv:2012.04035}, 2020.

\bibitem{vaswani2017attention}
Ashish Vaswani, Noam Shazeer, Niki Parmar, Jakob Uszkoreit, Llion Jones, Aidan~N Gomez, {\L}ukasz Kaiser, and Illia Polosukhin.
\newblock Attention is all you need.
\newblock {\em Advances in neural information processing systems}, 30, 2017.

\bibitem{wang2023learning}
Limei Wang, Haoran Liu, Yi~Liu, Jerry Kurtin, and Shuiwang Ji.
\newblock Learning hierarchical protein representations via complete 3d graph networks.
\newblock In {\em International Conference on Learning Representations (ICLR)}, 2023.

\bibitem{wang2005pdbbind}
Renxiao Wang, Xueliang Fang, Yipin Lu, Chao-Yie Yang, and Shaomeng Wang.
\newblock The pdbbind database: methodologies and updates.
\newblock {\em Journal of medicinal chemistry}, 48(12):4111--4119, 2005.

\bibitem{watson2023novo}
Joseph~L Watson, David Juergens, Nathaniel~R Bennett, Brian~L Trippe, Jason Yim, Helen~E Eisenach, Woody Ahern, Andrew~J Borst, Robert~J Ragotte, Lukas~F Milles, et~al.
\newblock De novo design of protein structure and function with rfdiffusion.
\newblock {\em Nature}, 620(7976):1089--1100, 2023.

\bibitem{wu2022discovering}
Fang Wu, Siyuan Li, Lirong Wu, Stan~Z Li, Dragomir Radev, and Qiang Zhang.
\newblock Discovering the representation bottleneck of graph neural networks from multi-order interactions.
\newblock {\em arXiv preprint arXiv:2205.07266}, 2022.

\bibitem{xu2022geodiff}
Minkai Xu, Lantao Yu, Yang Song, Chence Shi, Stefano Ermon, and Jian Tang.
\newblock Geodiff: A geometric diffusion model for molecular conformation generation.
\newblock {\em arXiv preprint arXiv:2203.02923}, 2022.

\bibitem{yu2023unified}
Qiying Yu, Yudi Zhang, Yuyan Ni, Shikun Feng, Yanyan Lan, Hao Zhou, and Jingjing Liu.
\newblock Unified molecular modeling via modality blending.
\newblock {\em arXiv preprint arXiv:2307.06235}, 2023.

\bibitem{zaidi2022pre}
Sheheryar Zaidi, Michael Schaarschmidt, James Martens, Hyunjik Kim, Yee~Whye Teh, Alvaro Sanchez-Gonzalez, Peter Battaglia, Razvan Pascanu, and Jonathan Godwin.
\newblock Pre-training via denoising for molecular property prediction.
\newblock {\em arXiv preprint arXiv:2206.00133}, 2022.

\bibitem{zhang2022protein}
Zuobai Zhang, Minghao Xu, Arian Jamasb, Vijil Chenthamarakshan, Aurelie Lozano, Payel Das, and Jian Tang.
\newblock Protein representation learning by geometric structure pretraining.
\newblock {\em arXiv preprint arXiv:2203.06125}, 2022.

\bibitem{zhou2023uni}
Gengmo Zhou, Zhifeng Gao, Qiankun Ding, Hang Zheng, Hongteng Xu, Zhewei Wei, Linfeng Zhang, and Guolin Ke.
\newblock Uni-mol: a universal 3d molecular representation learning framework, 2023.

\end{thebibliography}


\begin{thebibliography}{10}

\bibitem{Abraham2024}
M.~Abraham, A.~Alekseenko, V.~Basov, C.~Bergh, E.~Briand, A.~Brown, M.~Doijade, G.~Fiorin, S.~Fleischmann, S.~Gorelov, G.~Gouaillardet, A.~Grey, M.~E. Irrgang, F.~Jalalypour, J.~Jordan, C.~Kutzner, J.~A. Lemkul, M.~Lundborg, P.~Merz, and E.~Lindahl.
\newblock {\em GROMACS 2024.2 Manual (2024.2)}.
\newblock Zenodo, https://doi.org/10.5281/zenodo.11148638, 2024.

\bibitem{bussi2007canonical}
Giovanni Bussi, Davide Donadio, and Michele Parrinello.
\newblock Canonical sampling through velocity rescaling.
\newblock {\em The Journal of chemical physics}, 126(1), 2007.

\bibitem{case2021amber}
David~A Case, H~Metin Aktulga, Kellon Belfon, Ido Ben-Shalom, Scott~R Brozell, David~S Cerutti, Thomas~E Cheatham~III, Vin{\'\i}cius Wilian~D Cruzeiro, Tom~A Darden, Robert~E Duke, et~al.
\newblock {\em Amber 2021}.
\newblock University of California, San Francisco, 2021.

\bibitem{feng2023fractional}
Shikun Feng, Yuyan Ni, Yanyan Lan, Zhi-Ming Ma, and Wei-Ying Ma.
\newblock Fractional denoising for 3d molecular pre-training.
\newblock In {\em International Conference on Machine Learning}, pages 9938--9961. PMLR, 2023.

\bibitem{friesner2004glide}
Richard~A Friesner, Jay~L Banks, Robert~B Murphy, Thomas~A Halgren, Jasna~J Klicic, Daniel~T Mainz, Matthew~P Repasky, Eric~H Knoll, Mee Shelley, Jason~K Perry, et~al.
\newblock Glide: a new approach for rapid, accurate docking and scoring. 1. method and assessment of docking accuracy.
\newblock {\em Journal of medicinal chemistry}, 47(7):1739--1749, 2004.

\bibitem{Frisch2016}
M.~J. Frisch, G.~W. Trucks, H.~B. Schlegel, G.~E. Scuseria, M.~A. Robb, J.~R. Cheeseman, G.~Scalmani, V.~Barone, G.~A. Petersson, H.~Nakatsuji, X.~Li, M.~Caricato, A.~V. Marenich, J.~Bloino, B.~G. Janesko, R.~Gomperts, B.~Mennucci, H.~P. Hratchian, J.~V. Ortiz, A.~F. Izmaylov, J.~L. Sonnenberg, F.~Williams, Ding, F.~Lipparini, F.~Egidi, J.~Goings, B.~Peng, A.~Petrone, T.~Henderson, D.~Ranasinghe, V.~G. Zakrzewski, J.~Gao, N.~Rega, G.~Zheng, W.~Liang, M.~Hada, M.~Ehara, K.~Toyota, R.~Fukuda, J.~Hasegawa, M.~Ishida, T.~Nakajima, Y.~Honda, O.~Kitao, H.~Nakai, T.~Vreven, K.~Throssell, J.~A. Montgomery, J.~E. Peralta, F.~Ogliaro, M.~J. Bearpark, J.~J. Heyd, E.~N. Brothers, K.~N. Kudin, V.~N. Staroverov, T.~A. Keith, R.~Kobayashi, J.~Normand, K.~Raghavachari, A.~P. Rendell, J.~C. Burant, S.~S. Iyengar, J.~Tomasi, M.~Cossi, J.~M. Millam, M.~Klene, C.~Adamo, R.~Cammi, J.~W. Ochterski, R.~L. Martin, K.~Morokuma, O.~Farkas, J.~B. Foresman, and D.~J. Fox.
\newblock {\em Gaussian 09 Rev. D.01}.
\newblock Gaussian Inc., Wallingford, 2016.

\bibitem{gao2023self}
Bowen Gao, Yinjun Jia, Yuanle Mo, Yuyan Ni, Weiying Ma, Zhiming Ma, and Yanyan Lan.
\newblock Self-supervised pocket pretraining via protein fragment-surroundings alignment.
\newblock {\em arXiv preprint arXiv:2310.07229}, 2023.

\bibitem{godwin2021simple}
Jonathan Godwin, Michael Schaarschmidt, Alexander~L Gaunt, Alvaro Sanchez-Gonzalez, Yulia Rubanova, Petar Veli{\v{c}}kovi{\'c}, James Kirkpatrick, and Peter Battaglia.
\newblock Simple gnn regularisation for 3d molecular property prediction and beyond.
\newblock In {\em International Conference on Learning Representations}, 2021.

\bibitem{jiao2023energy}
Rui Jiao, Jiaqi Han, Wenbing Huang, Yu~Rong, and Yang Liu.
\newblock Energy-motivated equivariant pretraining for 3d molecular graphs.
\newblock In {\em Proceedings of the AAAI Conference on Artificial Intelligence}, volume~37, pages 8096--8104, 2023.

\bibitem{jin2023unsupervised}
Wengong Jin, Siranush Sarkizova, Xun Chen, Nir Hacohen, and Caroline Uhler.
\newblock Unsupervised protein-ligand binding energy prediction via neural euler's rotation equation.
\newblock {\em arXiv preprint arXiv:2301.10814}, 2023.

\bibitem{jing2021equivariant}
Bowen Jing, Stephan Eismann, Pratham~N Soni, and Ron~O Dror.
\newblock Equivariant graph neural networks for 3d macromolecular structure.
\newblock {\em arXiv preprint arXiv:2106.03843}, 2021.

\bibitem{kawata2001particle}
Masaaki Kawata and Umpei Nagashima.
\newblock Particle mesh ewald method for three-dimensional systems with two-dimensional periodicity.
\newblock {\em Chemical Physics Letters}, 340(1-2):165--172, 2001.

\bibitem{king2021recent}
Edward King, Erick Aitchison, Han Li, and Ray Luo.
\newblock Recent developments in free energy calculations for drug discovery.
\newblock {\em Frontiers in Molecular Biosciences}, 8:712085, 2021.

\bibitem{leach2022denoising}
Adam Leach, Sebastian~M Schmon, Matteo~T. Degiacomi, and Chris~G. Willcocks.
\newblock Denoising diffusion probabilistic models on {SO}(3) for rotational alignment.
\newblock In {\em ICLR 2022 Workshop on Geometrical and Topological Representation Learning}, 2022.

\bibitem{xFormers2022}
Benjamin Lefaudeux, Francisco Massa, Diana Liskovich, Wenhan Xiong, Vittorio Caggiano, Sean Naren, Min Xu, Jieru Hu, Marta Tintore, Susan Zhang, Patrick Labatut, and Daniel Haziza.
\newblock xformers: A modular and hackable transformer modelling library.
\newblock \url{https://github.com/facebookresearch/xformers}, 2022.

\bibitem{liao2022equiformer}
Yi-Lun Liao and Tess Smidt.
\newblock Equiformer: Equivariant graph attention transformer for 3d atomistic graphs.
\newblock {\em arXiv preprint arXiv:2206.11990}, 2022.

\bibitem{lin2022language}
Zeming Lin, Halil Akin, Roshan Rao, Brian Hie, Zhongkai Zhu, Wenting Lu, Nikita Smetanin, Allan dos Santos~Costa, Maryam Fazel-Zarandi, Tom Sercu, Sal Candido, et~al.
\newblock Language models of protein sequences at the scale of evolution enable accurate structure prediction.
\newblock {\em bioRxiv}, 2022.

\bibitem{lu2024comprehensive}
Tian Lu.
\newblock A comprehensive electron wavefunction analysis toolbox for chemists, multiwfn.
\newblock {\em The Journal of Chemical Physics}, 161(8), 2024.

\bibitem{Lu2024}
Tian Lu.
\newblock Sobtop (version 1.0), 2024.
\newblock Accessed: June 18, 2024.

\bibitem{maier2015ff14sb}
James~A Maier, Carmenza Martinez, Koushik Kasavajhala, Lauren Wickstrom, Kevin~E Hauser, and Carlos Simmerling.
\newblock ff14sb: improving the accuracy of protein side chain and backbone parameters from ff99sb.
\newblock {\em Journal of chemical theory and computation}, 11(8):3696--3713, 2015.

\bibitem{parrinello1981polymorphic}
Michele Parrinello and Aneesur Rahman.
\newblock Polymorphic transitions in single crystals: A new molecular dynamics method.
\newblock {\em Journal of Applied physics}, 52(12):7182--7190, 1981.

\bibitem{paszke2019pytorch}
Adam Paszke, Sam Gross, Francisco Massa, Adam Lerer, James Bradbury, Gregory Chanan, Trevor Killeen, Zeming Lin, Natalia Gimelshein, Luca Antiga, et~al.
\newblock Pytorch: An imperative style, high-performance deep learning library.
\newblock {\em Advances in neural information processing systems}, 32, 2019.

\bibitem{song2019generative}
Yang Song and Stefano Ermon.
\newblock Generative modeling by estimating gradients of the data distribution.
\newblock {\em Advances in neural information processing systems}, 32, 2019.

\bibitem{soper2008classical}
Davison~E Soper.
\newblock {\em Classical field theory}.
\newblock Courier Dover Publications, 2008.

\bibitem{tholke2022torchmd}
Philipp Th{\"o}lke and Gianni De~Fabritiis.
\newblock Torchmd-net: equivariant transformers for neural network based molecular potentials.
\newblock {\em arXiv preprint arXiv:2202.02541}, 2022.

\bibitem{townshend2020atom3d}
Raphael~JL Townshend, Martin V{\"o}gele, Patricia Suriana, Alexander Derry, Alexander Powers, Yianni Laloudakis, Sidhika Balachandar, Bowen Jing, Brandon Anderson, Stephan Eismann, et~al.
\newblock Atom3d: Tasks on molecules in three dimensions.
\newblock {\em arXiv preprint arXiv:2012.04035}, 2020.

\bibitem{valdes2021gmx_mmpbsa}
Mario~S Vald{\'e}s-Tresanco, Mario~E Vald{\'e}s-Tresanco, Pedro~A Valiente, and Ernesto Moreno.
\newblock gmx\_mmpbsa: a new tool to perform end-state free energy calculations with gromacs.
\newblock {\em Journal of chemical theory and computation}, 17(10):6281--6291, 2021.

\bibitem{wang2004development}
Junmei Wang, Romain~M Wolf, James~W Caldwell, Peter~A Kollman, and David~A Case.
\newblock Development and testing of a general amber force field.
\newblock {\em Journal of computational chemistry}, 25(9):1157--1174, 2004.

\bibitem{xu2022geodiff}
Minkai Xu, Lantao Yu, Yang Song, Chence Shi, Stefano Ermon, and Jian Tang.
\newblock Geodiff: A geometric diffusion model for molecular conformation generation.
\newblock {\em arXiv preprint arXiv:2203.02923}, 2022.

\bibitem{yasir2024investigating}
Muhammad Yasir, Jinyoung Park, Eun-Taek Han, Jin-Hee Han, Won~Sun Park, and Wanjoo Chun.
\newblock Investigating the inhibitory potential of flavonoids against aldose reductase: insights from molecular docking, dynamics simulations, and gmx\_mmpbsa analysis.
\newblock {\em Current Issues in Molecular Biology}, 46(10):11503--11518, 2024.

\bibitem{zaidi2022pre}
Sheheryar Zaidi, Michael Schaarschmidt, James Martens, Hyunjik Kim, Yee~Whye Teh, Alvaro Sanchez-Gonzalez, Peter Battaglia, Razvan Pascanu, and Jonathan Godwin.
\newblock Pre-training via denoising for molecular property prediction.
\newblock {\em arXiv preprint arXiv:2206.00133}, 2022.

\bibitem{zhang2023pre}
Zuobai Zhang, Minghao Xu, Aurelie Lozano, Vijil Chenthamarakshan, Payel Das, and Jian Tang.
\newblock Pre-training protein encoder via siamese sequence-structure diffusion trajectory prediction.
\newblock In {\em Annual Conference on Neural Information Processing Systems}, 2023.

\bibitem{zhou2023uni}
Gengmo Zhou, Zhifeng Gao, Qiankun Ding, Hang Zheng, Hongteng Xu, Zhewei Wei, Linfeng Zhang, and Guolin Ke.
\newblock Uni-mol: a universal 3d molecular representation learning framework, 2023.

\end{thebibliography}




\clearpage
\appendix
\begin{bibunit}
\label{FirstAppendixPage}
\newpage
\renewcommand{\thesection}{\arabic{section}} 

\title{Supplementary Information: An Equivariant Pretrained Transformer for Unified 3D Molecular Representation Learning}
\renewcommand{\abstracttitle}{}
\begin{abstract}
\end{abstract}
\settitle
\vspace{-1in}

\startlist{toc}
\printlist{toc}{}{\section*{Contents}}

\listofsupfigures

\listofsuptables


\setcounter{page}{1}
\setcounter{figure}{0}
\renewcommand{\figurename}{Supplementary Fig}
\setcounter{table}{0}
\renewcommand{\tablename}{Supplementary Table}

\clearpage

\section{Supplementary Methods}

\subsection{Dataset Collection}

\subsubsection{Detailed Dataset Distribution}

We collection the 3D molecule datasets from small molecules and protein domain, as detailed in Table~\ref{tab:pretrain_set}. Moreover, for each time loading data from PDB, we further randomly extract a local scope with three sequencially continuous residues as a training sample. This random segmentation approach is applied as the data augmentation for proteins. 

\begin{suptable}[h]
\centering
\caption[Statistics of the structural datasets for pretraining]{Statistics of the structural datasets for pretraining.}
\begin{tabular}{llrr}
\toprule
Domain                          & Source                 & \# of entries 
 & \# of blocks \\ \midrule
\multirow{3}{*}{Small Molecule} & GEOM-QM9               & 430,201  & 4,222,196     \\
                                & GEOM-Drugs             & 1,465,181  &
                                38,222,235 \\
                                & PCQM4Mv2               & 3,378,606   &
                                51,090,224 \\ \midrule
\multirow{4}{*}{Protein}        & PDB                    & 599,699   & 155,200,297    \\
                                & PDBBind-PP             & 2,852     & 2,990,905    \\
                                & PDBBind-refined-set    & 5,316     & 2,438,652    \\
                                & PDBBind-v2020-other-PL & 14,127    &
                                7,089,084 \\ \bottomrule
\end{tabular}
\label{tab:pretrain_set}
\end{suptable}

\subsubsection{Vocabulary Construction}
\label{apd:vocab}
The vocabulary of atom types, block types and position indexes are detailed in Table~\ref{tab:vocab}.

\begin{suptable}[h]
\centering
\caption[Construction of the vocabulary]{Construction of the vocabulary of atom types, block types and position indexes.}
\begin{tabular}{lll}
\toprule
Vocabulary & Index       & Descriptions                                                                                                               \\ \midrule
Atom       & 0$\sim$2    & \textless{}pad\textgreater{}, \textless{}mask\textgreater{}, \textless{}global\textgreater{}                               \\
           & 3$\sim$120  & 118 elements                                                                                                               \\ \midrule
Block      & 0$\sim$3    & \textless{}pad\textgreater{}, \textless{}mask\textgreater{}, \textless{}unk\textgreater{}, \textless{}global\textgreater{} \\
           & 4$\sim$23   & 20 amino acids                                                                                                             \\
           & 24$\sim$141 & 118 elements (H is included for completeness)                                                                              \\ \midrule
Position   & 0$\sim$2    & \textless{}pad\textgreater{}, \textless{}mask\textgreater{}, \textless{}global\textgreater{} \\
           & 3$\sim$12   & position codes for atoms in protein, i.e. $\alpha, \beta$, etc.                                                          \\
           & 13          & \textless{}sml\textgreater for atoms in small molecules  \\ \bottomrule                                                               
\end{tabular}
\label{tab:vocab}
\end{suptable}

\subsection{Comparison on Denoising Strategies}

In this section, we describe the block-level denoised pretraining approach designed to incorporate the hierarchical information into our backbone model (denoted as $\varphi$ hereinafter). The algorithm leverages the concept of Denoising Score Matching (DSM,~\cite{song2019generative}) to enable the model to learn from perturbed data representations. Generally, the overview of DSM is outlined in Algorithm \ref{alg:pretrain}. The training process begins by sampling perturbed coordinates $\vec{\mZ}'$ from a predefined noise distribution parameterized by $\sigma$ (Line 2). The atom-level pseudo forces $\vec{\mF}'$ are then predicted by the model $\varphi$ to recover $\vec{\mZ}$ from $\vec{\mZ}'$ (Line 3), and finally used to compute the denoising loss $\gL$ (Line 4). The key points of DSM lie in the design of the perturbation mechanism and the corresponding loss function to align the predicted forces $\vec{\mF}'$ with the Denoising Force Field (DFF). In the following, we first introduce the simple atom-level denoising method, then extend the denoising targets from atoms to blocks, and finally apply additional rotations on blocks to better depict the geometric landscape.

\begin{algorithm}[h]
\small
\caption{Overview of Denoised Pretraining}\label{alg:pretrain}
\begin{algorithmic}[1]
\STATE \textbf{Input:} Original Coordinates $\vec{\mZ}$, Noise scale $\sigma$, Backbone model $\varphi$.
\STATE Sample perturbed coordinates $\vec{\mZ}'\sim p_\sigma(\vec{\mZ}'|\vec{\mZ}).$
\STATE Predict pseudo forces $\vec{\mF}'\gets \varphi(\vec{\mZ}').$
\STATE Acquire denoising loss $\gL(\vec{\mF}', \vec{\mZ}', \vec{\mZ})$
\STATE Minimize $\gL$
\end{algorithmic}
\end{algorithm}

\textbf{Atom-level Denoising.} The atom-level denoising process~\cite{zaidi2022pre, jiao2023energy} independently introduces Gaussian noise to each atom's coordinate as $\vepsilon_\mZ\sim\gN(0,\mI_{3N})$ rescaled by $\sigma_t$:
\begin{align}
    \vec{\mZ}' &= C(\vec{\mZ} + \sigma_t\vepsilon_\mZ),
\end{align}

where the operation $C(\vec{\mZ})=\vec{\mZ} -\sum \vec{\mZ}/N$ projects the noised sample to the mean-centered subspace to neutralize the translation introduced by the noises~\cite{xu2022geodiff}.



The training objective is to match the predicted $\vec{\mF}'$ with the denoising force field yielded by $\nabla_{\vec{\mZ}'}p_{\sigma_t}(\vec{\mZ}'|\vec{\mZ})$ as
\begin{align}
\label{eq:prac_atom_loss}
    \gL_{\text{atom}} &= \E_{\vepsilon_\mZ\sim\gN(0,\mI_{3N})}\Big[\|\vec{\mF}'-\frac{\vec{\mZ}' - \vec{\mZ}}{\sigma_t^2}\|_2^2\Big].
\end{align}

\textbf{Translation-only Block-level Denoising.} To conserve the intra-block geometry, we extend the atom-level denoising task into block-level by considering blocks as rigid bodies, and all the atoms within the same block are applied by the same noise. For simplification, we first define the operators $\mu_b:\sR^{3\times N}\rightarrow\sR^{3\times M}, g_b:\sR^{3\times M}\rightarrow\sR^{3\times N}$ denote the atom-to-block averaging and the block-to-atom duplication. In particular, we have
\begin{equation}
 \begin{cases}
\mu_b(\vec{\mZ})[:,m_i] = \frac{\sum_{m_j=m_i} \vec{\vz}_j}{\sum_{m_j=m_i} 1}, \\
g_b(\vec{\mZ}_b)[:,i] = \vec{\mZ}_b[:, m_i].
\end{cases}   
\end{equation}

Slightly different from the atom-level setting, we apply noises on the center of each block as $\vepsilon_{\mZ_b}\sim\gN(0,\mI_{3M})$. For noise scale $\sigma_t$, the perturbation and mean-centered projection are sequentially calculated as

\begin{align}
    \vec{\mZ}' &= C\big(\vec{\mZ} + \sigma_t g_b(\vepsilon_{\mZ_b})\big).
\end{align}

The training objective adapts Eq.~(\ref{eq:prac_atom_loss}) into block-level as
\begin{align}
\label{eq:prac_blockt_loss}
    \gL_{\text{block-T}} &= \E_{\vepsilon_{\mZ_b}\sim\gN(0,\mI_{3M})}\Big[\|\mu_b(\vec{\mF}')-\frac{\mu_b(\vec{\mZ}') - \vec{\mZ}_b}{\sigma_t^2}\|_2^2\Big].
\end{align}

\textbf{Complete Block-level Denoising. (Ours)} Simply reducing $\vec{\mF}'$ into $\mu_b(\vec{\mF}')$ in Eq.~(\ref{eq:prac_blockt_loss}) ignores the torques applied on the blocks. To complete this point, we design an additional rotation denoising task from the perspective of Euler's rotation equation~\cite{soper2008classical}, which is previously proved effective on complex binding tasks~\citep{jin2023unsupervised}.

To begin with, the torque on each block is aggregated as
\begin{align}
    \vec{\mM}'_b[:, m_i] &= \sum_{m_j=m_i} (\vec{\vz}_j - \vec{\mZ}_b[:,m_i])\times\vec{\vf}'_j, 
\end{align}
According to Euler's rotation equation, the time derivative of the angular momentum of each block is given by
\begin{align}
\label{eq: euler}
    \frac{d\vec{\mL}_b}{dt} = \vec{\mM}'_b = \mI_b \vec{\valpha}_b,
\end{align}
where $\mI_b\in\sR^{3\times 3\times M}$ represents the inertia matrix defined as
\begin{align}
\label{eq: iner}
\mI_b[:, :, m_i] &= \sum_{m_j=m_i}\Big(\|\vec{\vu}_j\|^2\mI - \vec{\vu}_j \vec{\vu}_j^\top\Big),\\ 
\vec{\vu}_j &= \vec{\vz}_j - \vec{\mZ}_b[:,m_i].
\end{align}

The angular acceleration $\vec{\valpha}_b\in\sR^{3\times M}$ can be calculated by combining Eq.~(\ref{eq: euler}-\ref{eq: iner}) as
\begin{align}
    \vec{\valpha}_b=\mI_b^{-1} \vec{\mM}'_b.
\end{align}

To design an objective on $\vec{\valpha}_b$, we additionally perturb blocks by random rotations $\vomega_b$ sampled from the isotropic Gaussian distribution $\mathcal{IG}_{SO(3)}(\sigma_r)$~\citep{leach2022denoising}. Specifically, each rotation $\vomega_b[:,m_i]\in\mathfrak{so}(3)$ is constructed as $\vomega=\theta\hat{\vomega}$, where $\hat{\vomega}$ is a uniformly sampled unit vector, and $\theta\in[0,\pi]$ is a rotation angle with density
\begin{align}
    f(\theta)=\frac{1-\cos{\theta}}{\pi}\sum_{l=0}^\infty(2l+1)e^{-l(l+1)\sigma_r^2}\frac{\sin{\big((l+1/2)\theta\big)}}{\sin{(\theta/2)}}.
\end{align}

And the corresponding rotation matrix $\mQ(\vomega)\in SO(3)$ is acquired by the exponential mapping on $\vomega=(\omega_x,\omega_y,\omega_z)^\top$:
\begin{align}
    \mQ(\vomega)=\exp{
    \begin{bmatrix}
        0 & -\omega_z & \omega_y \\
        \omega_z & 0 & -\omega_x \\
        -\omega_y & \omega_x & 0
    \end{bmatrix}.
    }
\end{align}

Overall, the perturbation scheme combining block-level translation and rotation is designed as

\begin{align}
    \vec{\mZ}_r &= \vec{\mZ} - g_b(\vec{\mZ}_b), \\
    \vec{\mZ}' &= C\Big(g_b(\vec{\mZ}_b + \sigma_t\vepsilon_{\mZ_b}) + \mQ_b\vec{\mZ}_r\Big).
\end{align}

The translation loss is defined in Eq.~(\ref{eq:prac_blockt_loss}), and the rotation loss is defined as
\begin{align}
    \gL_{\text{block-R}} = \E_{\vomega\sim\mathcal{IG}_{SO(3)}(\sigma_r)}\Big[\|\vec{\valpha}_b - \nabla_{\vomega}p(\vomega)\|_2^2\Big].
\end{align}

The complete block-level training objective is added as
\begin{align}
    \gL_{\text{block-C}} = \gL_{\text{block-T}} + \gL_{\text{block-R}}.
\end{align}

\subsection{Implementation Details}

\subsubsection{Baseline Implementations on Virtual Screening Benchmark}

We benchmark EPT with the conventional software Glide~\citep{friesner2004glide} and the dual-tower model ESM+Uni-Mol, which utilizes the representations of proteins and small molecules from ESM~\citep{lin2022language} and Uni-Mol~\cite{zhou2023uni}, respectively. To be more specific, for each data pair consisting of the protein sequence and the drug SMILES, we leverage the pretrained ESM model \texttt{esm2\_t33\_650M\_UR50D}\footnote{https://github.com/facebookresearch/esm} and the pretrained Uni-Mol model from \texttt{unimol-tools}\footnote{https://unimol.readthedocs.io/en/latest} to obtain their representations. During training, representations of the protein and the small molecule are concatenated as the input to a simple $\mathrm{FFN}$ layer, while the output dimension differs for the two evaluation tasks. For convenience, we denote the dimension of the concatenated representations as $h_{\mathrm{in}}$. The hyperparameters for the dual tower model are listed in Table~\ref{tab:hyper_tower}.

\begin{suptable}[H]
\centering
\caption[Hyperparameters for the dual-tower model]{Hyperparameters for the dual-tower model.}
\resizebox{0.8\linewidth}{!}{
\begin{tabular}{lccccccc}
\toprule
Name               & \multicolumn{1}{c}{lr} & \multicolumn{1}{c}{batch\_size} & \multicolumn{1}{c}{epoch} & \multicolumn{1}{c}{$h_{\mathrm{in}}$} & \multicolumn{1}{c}{$h_{\mathrm{ffn}}$} & \multicolumn{1}{c}{activation} \\ \midrule
ESM+Uni-Mol & $1.0\times 10^{-3}$    & 16                              & 50                                 & 1792                        & 512  & SiLU       \\
\bottomrule
\end{tabular}
\label{tab:hyper_tower}
}
\end{suptable}

\subsubsection{Hyperparameters for Pretraining}

We pretrain EPT on 8 NVIDIA Tesla A800 with hyperparameters in Table~\ref{tab:hyper_ept}.

\begin{suptable}[h]
\centering
\caption[Hyperparameters for constructing and training EPT]{Hyperparameters for constructing and training EPT.}
\resizebox{0.95\linewidth}{!}{
\begin{tabular}{lcccccccc}
\toprule
Name  & $h_\text{hidden}$ & $h_\text{ffn}$ & $h_\text{edge}$     & $h_\text{rbf}$      & $L$                   & $H$                 & $\delta_\text{max}$ & $\delta_\text{topo}$ \\ \midrule
Value & 512               & 512            & 64                  & 64                  & 6                     & 8                   & 10.0                & 1.6                  \\ \midrule\midrule
Name  & epoch             & scheduler      & lr                  & min\_lr             & $\sigma_t$ & $\sigma_r$ & max\_n\_vertex      &   max\_vertex\_per\_gpu               \\ \midrule
Value & 50                & cosine         & $1.0\times 10^{-4}$ & $1.0\times 10^{-5}$ & 0.04                  & 0.1                 & 5,000               &       10,000              \\
\bottomrule
\end{tabular}
}
\label{tab:hyper_ept}
\end{suptable}

\subsubsection{Hyperparameters for LBA Task}

We utilize the pretrained model as the encoder and additional apply an output head to predict the affinity. Specifically, we consider three types of output heads based on an MLP $\varphi_E$ as follows:
\begin{align}
    \varphi_\text{atom}(\mH^{(l)}) &= \sum_i \varphi_E(\vh_i^{(l)}),\\
    \varphi_\text{block}(\mH^{(l)}) &= \sum_{m_i} \varphi_E\Big(\sum_{m_j=m_i}\vh_j^{(l)})\Big),\\
    \varphi_\text{graph}(\mH^{(l)}) &= \varphi_E(\sum_i\vh_i^{(l)}).
\end{align}
Based on these heads, the hyperparameters for finetuning on LBA are provided in Table~\ref{tab:hyper_lba}, with top-k checkpoints averaged for evaluation.

\begin{suptable}[h]
\centering
\caption[Hyperparameters for finetuning on LBA]{Hyperparameters for finetuning on LBA.}
\resizebox{0.95\linewidth}{!}{
\begin{tabular}{lcccccccccc}
\toprule
Name         & lr                  & batch\_size & output\_head & label\_norm & epoch & save\_topk & factor & patience & min\_lr             & omit\_sml\_pos \\ \midrule
\multicolumn{11}{c}{Sequence Identity 30}                                                                                                                   \\ \midrule
EPT-Scratch  & $1.0\times 10^{-4}$ & 16          & graph        & std         & 5    & 3          & 0.8    & 5        & $1.0\times 10^{-7}$ & True           \\
EPT-Molecule & $1.0\times 10^{-5}$ & 16          & block        & std         & 15    & 5          & 0.8    & 5        & $1.0\times 10^{-7}$ & True           \\
EPT-Protein  & $1.0\times 10^{-5}$ & 8           & graph         & none         & 15    & 3          & 0.8    & 5        & $1.0\times 10^{-7}$ & False          \\
EPT-MultiDomain   & $1.0\times 10^{-4}$ & 16          & graph        & mad         & 15    & 5          & 0.8    & 5        & $1.0\times 10^{-7}$ & False          \\ \midrule\midrule
\multicolumn{11}{c}{Sequence Identity 60}                                                                                                                    \\ \midrule
EPT-Scratch  & $5.0\times 10^{-5}$ & 8          & block        & none         & 30    & 5          & 0.8    & 5        & $1.0\times 10^{-7}$ & True           \\
EPT-Molecule & $5.0\times 10^{-5}$ & 8          & atom        & mad        & 30    & 3          & 0.8    & 5        & $1.0\times 10^{-7}$ & True           \\
EPT-Protein  & $5.0\times 10^{-5}$ & 8           & atom         & none        & 30    & 5          & 0.8    & 5        & $1.0\times 10^{-7}$ & False          \\
EPT-MultiDomain   & $5.0\times 10^{-5}$ & 16          & atom         & none        & 30    & 5          & 0.8    & 5        & $1.0\times 10^{-7}$ & False        \\ \bottomrule 
\end{tabular}
}
\label{tab:hyper_lba}
\end{suptable}

\subsubsection{Hyperparameters for MPP Task}

Following previous studies~\cite{zaidi2022pre, feng2023fractional}, we utilize the noisy node technique~\cite{godwin2021simple} by adding $\gL_\text{block-C}$ as an auxiliary training objective, and the entire loss for finetuning on QM9 can be formulated as $\gL=\gL_\text{MAE}+\lambda \gL_\text{block-C}$, where $\lambda$ balances the weight of each term. We utilize the same hyperparameters for all 12 tasks, which are detailed in Table~\ref{tab:hyper_qm9}.

\begin{suptable}[h]
\centering
\caption[Hyperparameters for finetuning on MPP]{Hyperparameters for finetuning on MPP.}
\resizebox{0.95\linewidth}{!}{
\begin{tabular}{lccccccccccc}
\toprule
Name   & lr                  & batch\_size & $\sigma_t$ & $\sigma_r$ & $\lambda$ & epoch &  factor & patience & min\_lr             & omit\_sml\_pos \\ \midrule
EPT    & $5.0\times 10^{-5}$ & 64          & 0.04       & 0.1        & 0.1       & 1,000  & 0.8    & 15       & $1.0\times 10^{-7}$ & True           \\
EPT-10 & $5.0\times 10^{-5}$ & 64          & 0.04       & 0.05       & 0.1       & 1,000  & 0.8    & 15       & $1.0\times 10^{-7}$ & True       \\ \bottomrule   
\end{tabular}
\label{tab:hyper_qm9}
}
\end{suptable}

\subsubsection{Hyperparameters for MSP Task}

We use the split by sequence identity over 30\% provided by Atom3D~\citep{townshend2020atom3d}, and extract all residues within 6\AA~distance to the mutation point as the local view for input, where the distance between two residues is measured by the minimum distance between atom pairs. The hyperparameters for finetuning MSP are listed in Table~\ref{tab:hyper_msp}, with the last checkpoint among the top-k checkpoints for evaluation.

\begin{suptable}[H]
\centering
\caption[Hyperparameters for finetuning on MSP]{Hyperparameters for finetuning on MSP.}
\resizebox{0.8\linewidth}{!}{
\begin{tabular}{lcccccccc}
\toprule
Name               & \multicolumn{1}{c}{lr} & \multicolumn{1}{c}{batch\_size} & \multicolumn{1}{c}{epoch} & \multicolumn{1}{c}{save\_topk} & \multicolumn{1}{c}{factor} & \multicolumn{1}{c}{patience} & \multicolumn{1}{c}{min\_lr} \\ \midrule
EPT (w/o Pretrain) & $1.0\times 10^{-5}$    & 16                              & 10        & 3                         & 0.6                        & 5                            & $5.0\times 10^{-6}$         \\
EPT (w/ Pretrain)  & $1.0\times 10^{-5}$    & 16                               & 10          & 2                       & 0.6                        & 5                            & $5.0\times 10^{-6}$        \\ \bottomrule
\end{tabular}
\label{tab:hyper_msp}
}
\end{suptable}

\subsubsection{Hyperparameters for Virtual Screening}

For the ranking task, we use $\mathrm{negative\_rate}$ to adjust the fraction of negative data pairs during training. Additionally, if $\mathrm{local\_mask}$ is set to be true, then only attentions from neighbors within $\delta_{\mathrm{max}}$ of each node will be calculated, otherwise attention values from all neighbors will be aggregated. The hyperparameters for finetuning Docked-PDBBind are listed in Table~\ref{tab:hyper_docked_pdbbind}.

\begin{suptable}[H]
\centering
\caption[Hyperparameters for finetuning on Docked-PDBBind]{Hyperparameters for finetuning on Docked-PDBBind, where \texttt{pred} indicates the affinity prediction task and \texttt{rank} indicates the ranking task.}
\resizebox{0.8\linewidth}{!}{
\begin{tabular}{lcccccc}
\toprule
Name               & \multicolumn{1}{c}{lr} & \multicolumn{1}{c}{batch\_size} & \multicolumn{1}{c}{epoch} & \multicolumn{1}{c}{negative\_rate} & \multicolumn{1}{c}{local\_mask} \\ \midrule
EPT (\texttt{pred}) & $3.0\times 10^{-5}$    & 16                              & 50                                 & 0.0                        & True         \\
EPT (\texttt{rank})  & $3.0\times 10^{-5}$    & 16                               & 50                                 & 0.6                        & False        \\ \bottomrule
\end{tabular}
\label{tab:hyper_docked_pdbbind}
}
\end{suptable}

\subsection{Additional Analyses on Candidate Ligands}

\subsubsection{Molecular Dynamic Simulation}

The SARS-CoV-2 3CL\textsuperscript{pro}-ligand complexes were simulated using GROMACS 2024.2~\cite{Abraham2024}. SARS-CoV-2 3CL\textsuperscript{pro} was topologized in the Amber14sb force field~\cite{maier2015ff14sb}. AmberTools 21.12 and Sobtop were used to paramterize the ligands in the general Amber force field (GAFF)~\cite{Lu2024,case2021amber}. The Restrained Electrostatic Potential (RESP) method~\cite{wang2004development} which was calcaulated by Multiwfn~\cite{lu2024comprehensive} and quantum chemical calculation (B3LYP functional at 6-31G(d,p)) were carried out to assign the atomic charge of the ligands by using Gaussian~\cite{Frisch2016}.

The parameterized SARS-CoV-2 3CL\textsuperscript{pro}-inhibitor complex was placed in a cubic simulation box, ensuring a minimum distance of $1.0 \,\text{nm}$ between the complex and the box edges. The system was then solvated with TIP3P water molecules. Counter ions (\textit{e.g.}, Na\textsuperscript{+} or Cl\textsuperscript{-}) were added to neutralize the solvated system. Following solvation and ion addition, energy minimization was performed using the steepest descent algorithm to remove any unfavorable contacts.

Thermodynamic properties such as pressure, density, potential energy, and temperature of the systems were monitored to ensure adequate equilibration before the production run. The Particle Mesh Ewald (PME) method was used to calculate the long-range electrostatics method with a Fourier grid spacing of 1.2\,\AA~\cite{kawata2001particle}. The LINCS algorithm~\cite{parrinello1981polymorphic} was applied to constrain covalent bonds during the equilibration steps. A modified V-rescale~\cite{bussi2007canonical} thermostat and Parrinello-Rahman barostat were used for temperature and pressure coupling, respectively. Finally, 100\,\text{ns} of unrestrained production simulations were carried out for the systems at 310\,\text{K} and 1\,\text{bar} atmospheric pressure.

\subsubsection{Binding Free Energy Calculation}

The program MMPBSA 1.6.3~\cite{valdes2021gmx_mmpbsa} was developed to compute the end state-free energies of protein-ligand complexes from GROMACS MD trajectory data. Binding free energy predictions were made using an MM/PBSA approach from the MD simulation trajectories in explicit solvent, analyzing three components: the complex, receptor, and ligand~\cite{king2021recent}. In our calculation, the trajectory data from the last 10\,\text{ns} was utilized, comprising 40 complex frames. The binding free energy $\Delta G_{\text{binding}}$ of the lead compounds in complex with the protein was determined using the following equation:
\begin{align}
    \Delta G_{\text{binding}} = \Delta G_{\text{complex}} - (\Delta G_{\text{protein}} + \Delta G_{\text{ligand}}).
\end{align}
In this equation, $\Delta G_{\text{complex}}$ represents the energy of the lead compound-protein complexes, and $\Delta G_{\text{protein}}~\text{and}~\Delta G_{\text{ligand}}$ demonstrate the energy of proteins and ligands in an aqueous environment, respectively~\cite{yasir2024investigating}. Supplementary Fig.~\ref{fig:mem1} shows a comprehensive analysis of the ten candidates screened based on EPT, which also includes the $\Delta G_{\text{binding}}$ of these ten candidates.

\begin{supfigure}[H]
\centering
\includegraphics[width=0.8\linewidth]{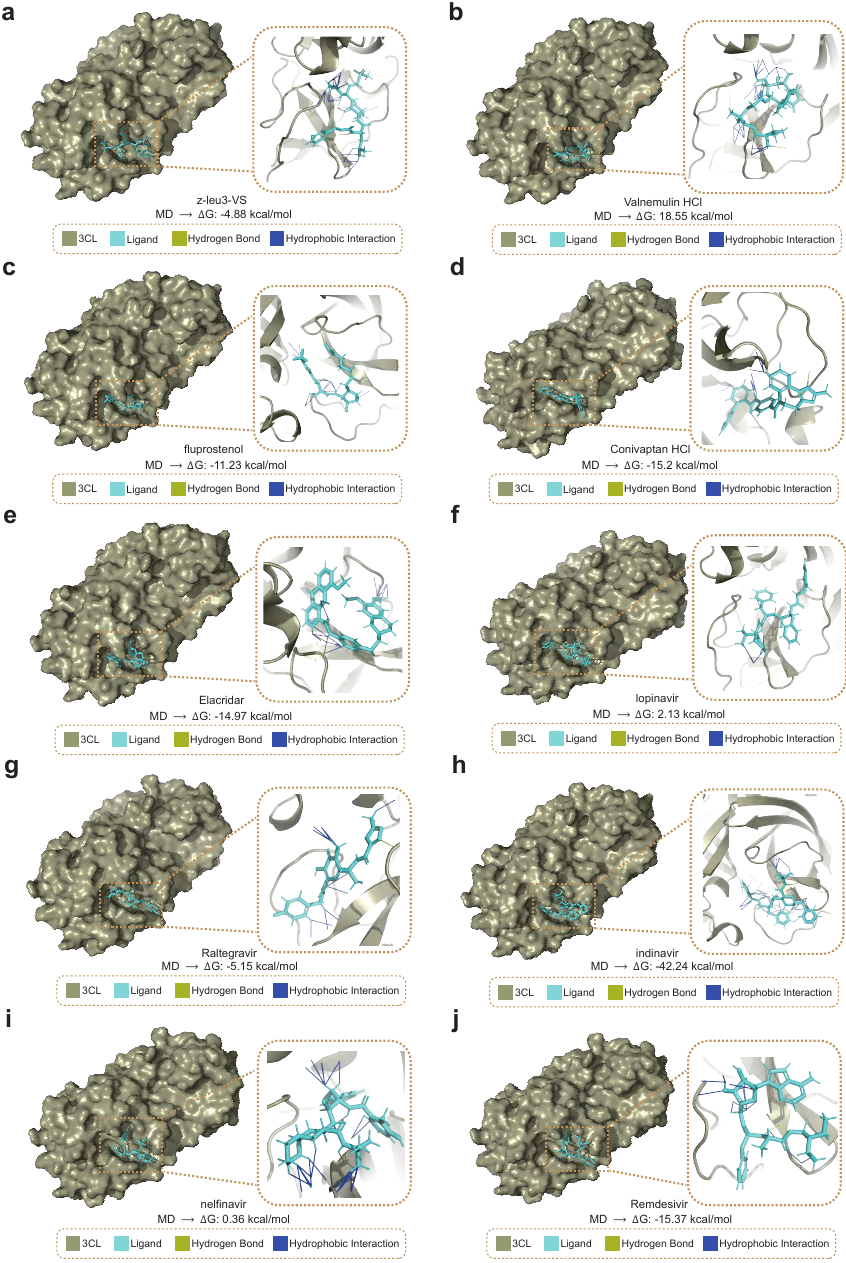}
\vskip -0.1in
\caption[A comprehensive analysis of the ten candidates screened based on EPT.]{\textbf{\ \textbar \ A comprehensive analysis of the ten candidates screened based on EPT.} \textbf{a-e}, Docking visualization and MD simulation results for the remaining molecules, excluding Ac-Leu-Leu-Nle-CHO, among the top six drugs ranked from the 1,944 FDA-approved drugs. \textbf{f-j}, Docking visualization and MD simulation results for the remaining molecules, excluding Saquinavir, among the top six drugs ranked from the 26 antiviral drugs. Note: The positive reference for the above ten candidates is also Leritrelvir ($\Delta G = -6.16\,\text{kcal/mol}$).}
\label{fig:mem1}
\end{supfigure}

\subsection{Raw Results}

\subsubsection{Results for LBA task}

Raw results for the Ligand Binding Affinity prediction (LBA) task are provided in Table~\ref{tab:lba}, where the baseline results are derived from ProtFSA~\cite{gao2023self}.

\begin{suptable*}[h]
\centering
\caption[Results on LBA]{The mean and standard deviations of 3 runs on the LBA dataset. The best results are in \textbf{bold} and the second best are \underline{underlined}.}
\resizebox{\linewidth}{!}{
\begin{tabular}{lcccccc}
\toprule
\multirow{2}{*}{Model}   & \multicolumn{3}{c}{Sequence Identity 30\%}                                        & \multicolumn{3}{c}{Sequence Identity 60\%}                                        \\ \cmidrule{2-4}\cmidrule{5-7}%

                         & RMSE$\downarrow$                      & Pearson$\uparrow$                    & Spearman$\uparrow$                  & RMSE$\downarrow$                       & Pearson$\uparrow$                    & Spearman$\uparrow$                   \\
\midrule
DeepDTA                  & $1.866\pm 0.080$          & $0.472\pm 0.022$          & $0.471\pm 0.024$          & $1.762\pm 0.261$          & $0.666\pm 0.012$          & $0.663\pm 0.015$          \\
B \& B  & $1.985\pm 0.006$          & $0.165\pm 0.006$          & $0.152\pm 0.024$          & $1.891\pm 0.004$          & $0.249\pm 0.006$          & $0.275\pm 0.008$          \\
TAPE                     & $1.890\pm 0.035$          & $0.338\pm 0.044$          & $0.286\pm 0.124$          & $1.633\pm 0.016$          & $0.568\pm 0.033$          & $0.571\pm 0.021$          \\
ProtTrans                & $1.544\pm 0.015$          & $0.438\pm 0.053$          & $0.434\pm 0.058$          & $1.641\pm 0.016$          & $0.595\pm 0.014$          & $0.588\pm 0.009$          \\ \midrule
MaSIF                    & $1.484\pm 0.018$          & $0.467\pm 0.020$          & $0.455\pm 0.014$          & $1.426\pm 0.017$          & $0.709\pm 0.008$          & $0.701\pm 0.001$          \\
IEConv                   & $1.554\pm 0.016$          & $0.414\pm 0.053$          & $0.428\pm 0.032$          & $1.473\pm 0.024$          & $0.667\pm 0.011$          & $0.675\pm 0.019$          \\
Holoprot-Full Surface    & $1.464\pm 0.006$          & $0.509\pm 0.002$          & $0.500\pm 0.005$          & $1.365\pm 0.038$          & $0.749\pm 0.014$          & $0.742\pm 0.011$          \\
Holoprot-Superpixel      & $1.491\pm 0.004$          & $0.491\pm 0.014$          & $0.482\pm 0.032$          & $1.416\pm 0.022$          & $0.724\pm 0.011$          & $0.715\pm 0.006$          \\
ProtNet-Amino Acid        & $1.455\pm 0.009$          & $0.536\pm 0.012$          & $0.526\pm 0.012$          & $1.397\pm 0.018$          & $0.741\pm 0.008$          & $0.734\pm 0.009$          \\
ProtNet-Backbone         & $1.458\pm 0.003$          & $0.546\pm 0.007$          & $0.550\pm 0.008$          & $1.349\pm 0.019$          & $0.764\pm 0.006$          & $0.759\pm 0.001$          \\
ProtNet-All-Atom         & $1.463\pm 0.001$          & $0.551\pm 0.005$          & $0.551\pm 0.008$          & $1.343\pm 0.025$          & $0.765\pm 0.009$          & $0.761\pm 0.003$          \\
Atom3D-3DCNN             & \textbf{$1.416\pm 0.021$} & \textbf{$0.550\pm 0.021$} & \textbf{$0.553\pm 0.009$} & \textbf{$1.621\pm 0.025$} & \textbf{$0.608\pm 0.020$} & \textbf{$0.615\pm 0.028$} \\
Atom3D-ENN               & \textbf{$1.568\pm 0.012$} & \textbf{$0.389\pm 0.024$} & \textbf{$0.408\pm 0.021$} & \textbf{$1.620\pm 0.049$} & \textbf{$0.623\pm 0.015$} & \textbf{$0.633\pm 0.021$} \\
Atom3D-GNN               & \textbf{$1.601\pm 0.048$} & \textbf{$0.545\pm 0.027$} & \textbf{$0.533\pm 0.033$} & \textbf{$1.408\pm 0.069$} & \textbf{$0.743\pm 0.022$} & \textbf{$0.743\pm 0.027$} \\ \midrule
EGNN-PLM                 & \textbf{$1.403\pm 0.010$} & \textbf{$0.565\pm 0.020$} & \textbf{$0.544\pm 0.010$} & \textbf{$1.559\pm 0.020$} & \textbf{$0.644\pm 0.020$} & \textbf{$0.646\pm 0.020$} \\
Uni-Mol         & \textbf{$1.520\pm 0.030$} & \textbf{$0.558\pm 0.000$} & \textbf{$0.540\pm 0.000$} & \textbf{$1.619\pm 0.040$} & \textbf{$0.645\pm 0.020$} & \textbf{$0.653\pm 0.020$} \\
ProFSA                   & \textbf{$1.377\pm 0.010$} & \underline{\textbf{$0.628\pm 0.010$}} & \underline{\textbf{$0.620\pm 0.010$}} & \textbf{$1.377\pm 0.010$} & \textbf{$0.764\pm 0.000$} & \textbf{$0.762\pm 0.010$} \\

\midrule
        EPT-Scratch & \textbf{$1.378\pm 0.026$} & \textbf{$0.604\pm 0.016$} & \textbf{$0.594\pm 0.017$} & \textbf{$1.277\pm 0.020$} & \textbf{$0.787\pm 0.008$} & \textbf{$0.785\pm 0.008$}  \\ 
        EPT-Molecule & \textbf{$1.336\pm 0.030$} & \textbf{$0.621\pm 0.016$} & \textbf{$0.602\pm 0.017$} & \textbf{$1.243\pm 0.022$} & \textbf{$0.802\pm 0.008$} & \underline{\textbf{$0.800\pm 0.009$}}  \\ 
        EPT-Protein & \underline{\textbf{$1.329\pm 0.013$}} & \underline{\textbf{$0.628\pm 0.004$}} & \textbf{$0.613\pm 0.004$} & \underline{\textbf{$1.235\pm 0.016$}} & \underline{\textbf{$0.804\pm 0.006$}} & \underline{\textbf{$0.800\pm 0.005$}}  \\ 
        EPT-MultiDomain & \textbf{$\mathbf{1.322\pm 0.025}$} & \textbf{$\mathbf{0.644\pm 0.016}$} & \textbf{$\mathbf{0.630\pm 0.014}$} & \textbf{$\mathbf{1.227\pm 0.013}$} & \textbf{$\mathbf{0.811\pm 0.003}$} & \textbf{$\mathbf{0.803\pm 0.004}$} \\ 
\bottomrule
\end{tabular}
  \label{tab:lba}
}
\end{suptable*}

\subsubsection{Results for MSP task}

Raw results for the Mutation Stability Prediction (MSP) task are provided in Table~\ref{tab:msp}, where the baseline results are derived from GVP~\cite{jing2021equivariant} and SiamDiff~\cite{zhang2023pre}.

\begin{suptable}[h]
\centering
\caption[Results on MSP]{Mean and standard deviation across 3 runs on mutation stability prediction (MSP). The best scores are marked in \textbf{bold} and the second best \underline{underlined}.}
\label{tab:msp}
\scalebox{0.8}{
\begin{tabular}{llcc}
\toprule
                              & Model & AUROC         \\ \midrule
\multirow{6}{*}{\begin{tabular}[c]{@{}l@{}}w/o \\ Pretrain\end{tabular}} & GCN                     & $0.621\pm 0.009$       \\
                              & Atom3D-CNN             & $0.574\pm 0.005$       \\
                              & Atom3D-ENN             & $0.574\pm 0.040$       \\
                              & GVP                    & $\underline{0.680\pm 0.015}$ \\
                              & GearNet-Edge           & $0.633\pm 0.067$       \\
                              & EPT (ours)             & $\bm{0.705\pm 0.025}$           \\ \hline
\multirow{4}{*}{\begin{tabular}[c]{@{}l@{}}w/ \\ Pretrain\end{tabular}} & Multiview Contrast &	$0.646\pm	0.006$ \\
 & DiffPreT &	$0.680\pm	0.018$ \\
 & SiamDiff &	$0.698\pm	0.020$  \\
                              & EPT (ours)             &         $\bm{0.720\pm 0.008}$    \\ \bottomrule
\end{tabular}}
\end{suptable}

\subsubsection{Results for Virtual Screening}

Raw results for the Virtual Screening task are provided in Table~\ref{tab:docked-pdbbind}, where the baseline results are derived from Glide~\cite{friesner2004glide}, ESM~\cite{lin2022language} and Uni-Mol~\cite{zhou2023uni}.

\begin{suptable*}[h]
\centering
\caption[Results on Docked PDBBind]{Results on 3 splits of Docked PDBBind. The best scores are marked in \textbf{bold} and the second best \underline{underlined}.}
\label{tab:docked-pdbbind}
\begin{tabular}{llcccc}
\hline
Split                     & Model       & |Pearson|$\uparrow$ & |Spearman|$\uparrow$ & Top-1 Acc$\uparrow$ & AvgRank$\downarrow$ \\ \hline
\multirow{3}{*}{id30}     & Glide       & 0.232               & 0.239                & 0.460       & 0.251               \\
                          & ESM+Uni-Mol & 0.201               & 0.228                & 0.202             & 0.371               \\
                          & EPT         & 0.422               & 0.430                & 0.454             & 0.249         \\ \hline
\multirow{3}{*}{id60}     & Glide       & 0.232               & 0.239                & 0.460       & 0.251               \\
                          & ESM+Uni-Mol & 0.180               & 0.163                & 0.236             & 0.347               \\
                          & EPT         & 0.445         & 0.451          & 0.454             & 0.258               \\ \hline
\multirow{3}{*}{scaffold} & Glide       & 0.232               & 0.239                & 0.460       & 0.251               \\
                          & ESM+Uni-Mol & 0.247               & 0.256                & 0.230             & 0.338               \\
                          & EPT         & 0.506      & 0.510       & 0.470    & 0.238      \\ \hline
\end{tabular}%
\end{suptable*}

\subsection{Memory Efficiency}

\label{apd:attn}
\subsubsection{Efficient Attention Mechanism}

Consider the attention mechanism introduced in Eq.~(\ref{eq:attn_begin}-\ref{eq:attn_end}), the query, key matrix $\mQ_s, \mK_s$ and the concatenated value matrix $\mV_s$ share the shape of $\sR^{B\times N_\text{max}\times S\times 4h_s}$, where $B, N_\text{max}, S, h_s$ denote the batch size, the maximum number of atoms, the number of heads and the size of each head's hidden state. Such shape consistency enables the usage of previous memory efficient techniques~\cite{xFormers2022} that provide the interface of attention biases for $\mD$ and $\mR$. We provide the pseudo codes in PyTorch~\cite{paszke2019pytorch} style as follows.

\newpage
\begin{lstlisting}[language=Python]
from xformers.ops import memory_efficient_attention

def equivariant_memory_efficient_self_attention(H_in, V_in, D, R, mask):

    ```
    Params:
        H_in: B * N_max * h
        V_in: B * N_max * 3 * h
        D: B * N_max * N_max
        R: B * N_max * N_max
        mask: B * N_max
    Returns:
        H_out: B * N_max * h
        V_out: B * N_max * 3 * h
    ```
    # Eq. (6)
    Query_s = linear_scalar_Q(H_in).view(B, N_max, S, h_s * 4)
    Key_s = linear_scalar_K(H_in).view(B, N_max, S, h_s * 4)
    Value_s_scalar = linear_scalar_K(H_in).view(B, N_max, S, h_s)
    Value_s_vector = linear_scalar_K(V_in).view(B, N_max, 3, S, h_s)
    # B * N_max * S * 3h_s
    Value_s_vector = Value_s_vector.transpose(2,3).flatten(start_dim=-2) 
    # B * N_max * S * 4h_s
    Value_s = cat([Value_s_scalar, Value_s_vector], dim=-1) 
    # Eq. (7)
    bias = R - D
    bias = bias.masked_fill(mask.unsqueeze(1).unsqueeze(2) == 0, float("inf"))
    bias = bias.expand(-1, S, -1, -1)
    HV = memory_efficient_attention(
        query = Query_s, 
        key = Key_s, 
        value = Value_s,
        attn_bias = bias
    )
    H_s = HV[..., :h_s]
    V_s = HV[..., h_s:].view(B, N_max, S, 3, h_s).transpose(2,3) 
    # Eq. (9)
    H_out = linear_scalar_O(H_s.view(B, N_max, h))
    V_out = linear_vector_O(V_s.view(B, N_max, 3, h))
    return H_out, V_out
    
\end{lstlisting}

\subsubsection{Comparison on Transformer-based Backbones}

We further compared the GPU memory consumption of our model, a 6-layer, 512-hidden EPT, with two previous Transformer-based backbones: the 6-layer, 512-hidden TorchMD-Net~\citep{tholke2022torchmd} and the 6-layer, 128-hidden, 3-degree Equiformer~\citep{liao2022equiformer}, which have 31M, 19M, and 18M parameters, respectively. Our tests measured memory usage of one forward step on point clouds with 32 to 2048 nodes, sampled uniformly within a sphere of radius $\sqrt[3]{N}$, and connected by edges within a 4.0 cutoff distance. Supplementary Fig.~\ref{fig:mem} illustrates that EPT is consistently more memory-efficient across various node counts. This efficiency enables our model to effectively process large-scale point clouds, facilitating the study of expansive molecular systems.

\begin{supfigure}[H]
\centering
\includegraphics[width=0.6\linewidth]{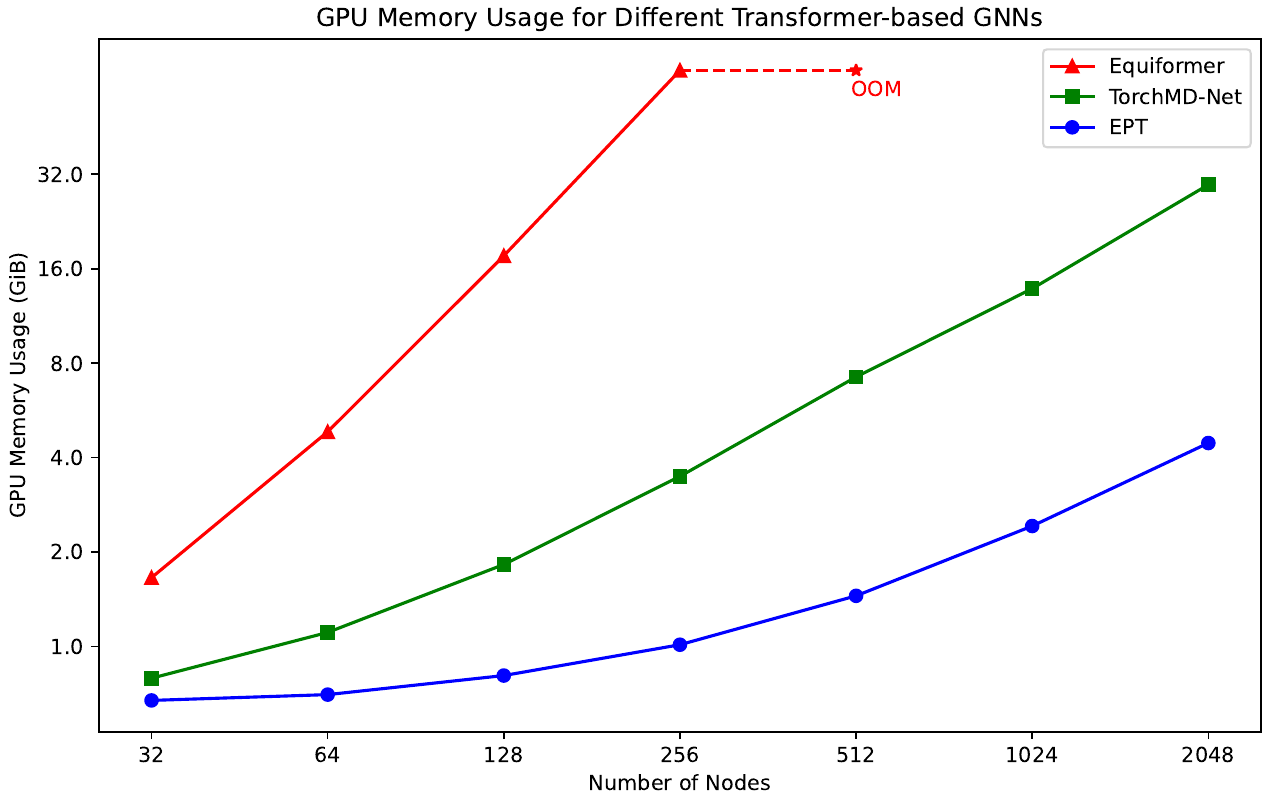}
\vskip -0.1in
\caption[GPU Memory Usage for Transformer-based GNNs]{\textbf{\ \textbar \ GPU Memory Usage for Transformer-based GNNs.} GPU memory usage comparison of three Transformer-based backbones as the number of nodes increases. EPT demonstrates a more memory-efficient scaling behavior compared to TorchMD and EquiFormer. EquiFormer encounters an Out-of-Memory (OOM) error at 512 nodes on a NVIDIA Tesla A800 with 80G GPU memory.}
\label{fig:mem}
\end{supfigure}

\putbib[ref]
\label{LastAppendixPage}
\end{bibunit}

\end{document}